\documentclass[11pt]{article}

\usepackage[final]{acl}
\usepackage{times}
\usepackage{latexsym}

\usepackage[T1]{fontenc}
\usepackage[utf8]{inputenc}

\usepackage{microtype}

\usepackage{inconsolata}

\usepackage{graphicx}
\usepackage{url}
\usepackage{soul}
\usepackage{amsmath}
\usepackage{amssymb}
\usepackage{mathtools}
\usepackage{amsthm}
\usepackage{booktabs}
\usepackage{multirow}
\usepackage{array}
\usepackage[table]{xcolor}
\usepackage{graphicx}
\usepackage{algorithm}
\usepackage{algpseudocode}
\usepackage{float}
\definecolor{lightgray}{RGB}{230,230,230}
\definecolor{lightblue}{RGB}{220,235,247}
\usepackage{adjustbox}

\usepackage{xcolor}
\usepackage{enumitem}
\usepackage[most]{tcolorbox}

\tcbset{
  promptboxstyle/.style={
    enhanced,
    breakable,
    sharp corners,
    boxrule=0.5pt,
    left=1mm,
    right=1mm,
    top=0.6mm,
    bottom=0.6mm,
    before upper=\setlength{\parindent}{0pt},
    fonttitle=\bfseries,
    fontupper=\footnotesize
  }
}

\tcbuselibrary{skins,breakable}
\newcommand{\tagbase}[2]{\texttt{\small\textcolor{#1}{#2}}}

\newcommand{\tagReason}[1]
{\tagbase{orange!85!black}{#1}}

\newcommand{\tagAnswer}[1]
{\tagbase{green!55!black}{#1}}

\newcommand{\tagSearch}[1]
{\tagbase{red!75!black}{#1}}

\newcommand{\tagEv}[1]
{\tagbase{purple!70!black}{#1}}

\newcommand{\tagCap}[1]
{\tagbase{blue!70!black}{#1}}

\title{Learning to Search: A Decision-Based Agent for Knowledge-Based Visual Question Answering}

\author{
 \textbf{Zhuohong Chen\textsuperscript{1}$^\dagger$},
 \textbf{Zhenxian Wu\textsuperscript{1}$^{\dagger}$},
 \textbf{Yunyao Yu\textsuperscript{1}$^{\dagger}$},
 \textbf{Hangrui Xu\textsuperscript{1}},
 \textbf{Zirui Liao\textsuperscript{1}},
 \textbf{Zhifang Liu}\textsuperscript{1},\\
 \textbf{Xiangwen Deng}\textsuperscript{2},
 \textbf{Pen Jiao}\textsuperscript{1},
 \textbf{Haoqian Wang\textsuperscript{1}\thanks{
   $\dagger$ Equal contribution \quad 
   Co-first authors: zhuohong24@mails.tsinghua.edu.cn \\
   $^*$ Corresponding authors: wanghaoqian@tsinghua.edu.cn
 }},
\\
 \textsuperscript{1}Tsinghua University,
 \textsuperscript{2}University of Arizona,
}

\begin{document}
\maketitle
\begin{figure*}[t]
    \centering
    \includegraphics[width=\textwidth]{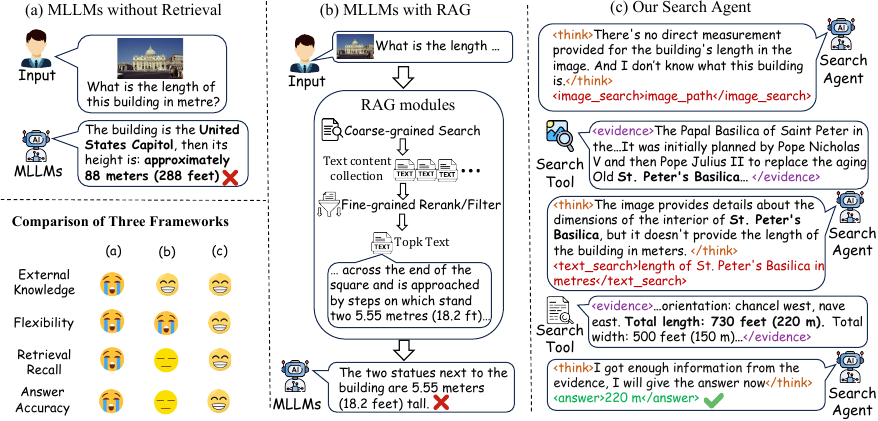}
    \caption{\textbf{Comparison of three paradigms for KB-VQA: MLLMs without retrieval, MLLMs with RAG, and our search agent.}
\textbf{(a) MLLMs without Retrieval} lack access to external knowledge and often fail on long-tail queries.
\textbf{(b) MLLMs with RAG} introduce external context but may suffer from noisy or mismatched retrieval due to fixed pipelines.
\textbf{(c) Our Search Agent} formulates KB-VQA as a multi-step decision process, enabling adaptive tool usage and structured evidence accumulation.}
    \label{fig:teaser}
\end{figure*}

\begin{abstract}
Knowledge-based visual question answering (KB-VQA) requires vision-language models to understand images and use external knowledge, especially for rare entities and long-tail facts. Most existing retrieval-augmented generation (RAG) methods adopt a fixed pipeline that sequentially retrieves information, filters it, and then produces an answer. Such a design makes it difficult to adapt to diverse question types. Moreover, it separates retrieval from reasoning, making it hard for the model to decide when to search, how to refine queries, or when to stop. As a result, the retrieved evidence is often poorly aligned with the question.
To address these limitations, we reformulate KB-VQA as a search-agent problem and model the solving process as a multi-step decision-making procedure. At each step, the agent selects one of four actions-Answer, Image Retrieval, Text Retrieval, and Caption-based on its current information state. We further design an automated pipeline to collect multi-step trajectories that record the agent’s reasoning process, tool usage, and intermediate decisions. These trajectories are then used as supervision for fine-tuning.
Experiments on InfoSeek and E-VQA demonstrate that our method achieves state-of-the-art performance, consistently outperforming prior baselines and confirming the effectiveness of our framework.
\end{abstract}

\begin{figure*}[t]
    \centering
    \includegraphics[width=\textwidth]{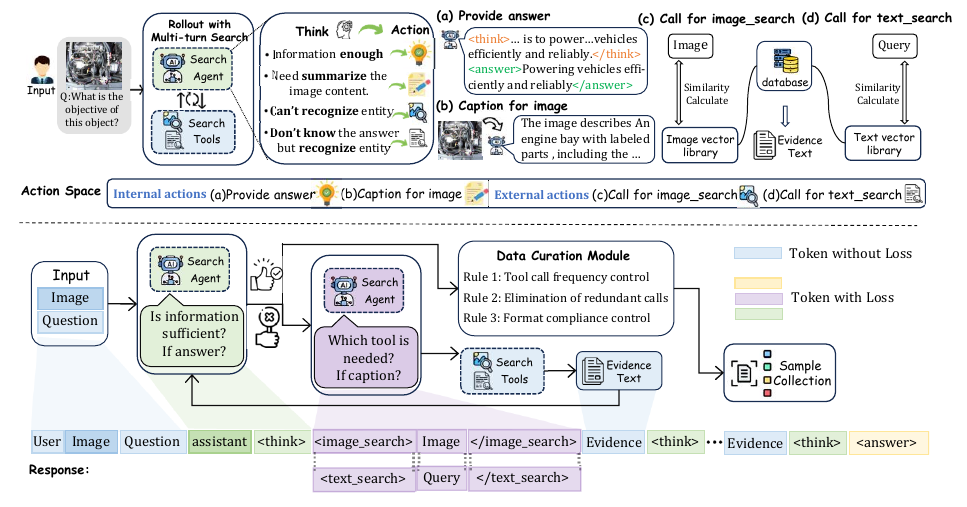}
    \caption{\textbf{Overview of the proposed DBAgent framework and trajectory-based training pipeline.}
\textbf{Top:} Inference-time decision-making, where the agent dynamically selects among multiple actions (e.g., image retrieval, text retrieval, captioning, and answering) based on the current information state.
\textbf{Below:} Construction of multi-step reasoning trajectories, data curation rules, and linearization into structured token sequences for SFT.}

    \label{fig:pipeline}
\end{figure*}

\section{Introduction}

In recent years, Vision--Language Models (VLMs) have achieved remarkable progress on multimodal tasks such as Visual Question Answering (VQA), demonstrating strong capabilities in cross-modal understanding and generation in recent benchmarks\cite{Tanaka2025VDocRAGRG,Xiao2024VideoQAIT}. 
However, when questions involve fine-grained entities or long-tail encyclopedic knowledge, relying solely on the knowledge internalized in model parameters is often insufficient to support accurate reasoning, leading to an overall degradation in performance\cite{Long2025RetrievalAugmentedVQ,Liu2026OVIBenchBO}.
This has led to the emergence of research in knowledge-based visual question answering (KB-VQA), which requires not only understanding visual content but also dynamically acquiring relevant information from external knowledge sources.

Retrieval-Augmented Generation (RAG) has become a dominant paradigm in KB-VQA. 
These methods retrieve candidate knowledge from large external corpora, such as Wikipedia, and inject the retrieved content into the model context to expand its knowledge coverage\cite{Choi2025MultimodalIR}. 
Recent studies have explored various directions, including multimodal query construction, cross-modal retrieval, and post-retrieval filtering, which together have led to noticeable improvements in KB-VQA performance\cite{Hong2025KnowledgebasedVQ,Deng2025EnablingCP}.

Despite these advances, most existing approaches still follow a modular and static pipeline design. 
This design implicitly assumes that all questions should follow the same retrieval and reasoning process. 
However, KB-VQA questions differ significantly in their dependence on external knowledge. 
While some questions can be answered directly without external retrieval, others require reflective retrieval to gather sufficient evidence. As a result, a single retrieval step with paragraph-level reranking is often insufficient.
By treating retrieval and reasoning as a fixed pipeline, existing methods struggle to explicitly track and exploit the current information state during inference, making it difficult to decide when to stop retrieval or how to adapt queries based on accumulated evidence, which often results in poor semantic alignment between retrieved content and the question.

Based on these observations, we argue that retrieval and reasoning in KB-VQA should be viewed as a dynamic decision-making process rather than a fixed pipeline. We therefore reformulate KB-VQA under a \emph{search agent} paradigm, where the model no longer passively consumes predefined retrieval results. 
Instead, it acts as an agent that decides whether the current state is sufficient, which external tool can provide the missing support, and when to terminate the search, conditioned on the accumulated multimodal evidence.
Specifically, we formalize the entire problem-solving process as a multi-step decision sequence with a structured multimodal action space:
\textbf{Answer} for generating the final answer;
\textbf{Image Retrieval} for performing image-based similarity search and returning top-ranked matched images together with their associated documents as candidate evidence;
\textbf{Text Retrieval} for retrieving relevant textual passages given a textual query; and
\textbf{Caption} for converting visual content into task-oriented descriptive text to facilitate evidence retrieval and answer generation.
During inference, the agent continuously analyzes the current multimodal information state and selects the next action at each step until evidence is gathered to support a final answer. 
To equip the model with both multimodal understanding and effective tool-selection abilities, we further design an automated multi-stage framework for constructing reasoning--retrieval trajectories. 
Based on this framework, we generate standardized trajectory data covering diverse KB-VQA scenarios, including different action types, retrieval depths, and evidence accumulation patterns. 
Each trajectory annotates the complete decision path from the initial input to the final answer. 
Using these trajectories, we perform supervised fine-tuning (SFT) to teach the model to select appropriate actions under different information states.
Experiments on two long-tail KB-VQA benchmarks, InfoSeek\cite{Chen2023CanPV} and E-VQA\cite{Mensink2023EncyclopedicVV}, demonstrate that our method outperforms existing approaches in both accuracy and reasoning stability.
Our main contributions can be summarized as follows:
\begin{itemize}
    \item We study KB-VQA from an agent perspective, identify the limitations of modular RAG pipelines in dynamic planning and state awareness, and propose a decision-based search-agent framework that decides when to answer, when and how to retrieve, and when to stop.
    \item We construct multi-step decision trajectory data on InfoSeek and E-VQA, covering diverse evidence acquisition patterns and train the model through supervised fine-tuning to jointly learn reasoning and tool usage.
    \item Our approach achieves significant improvements over prior methods on two mainstream KB-VQA benchmarks, InfoSeek and E-VQA, demonstrating the effectiveness of dynamic planning-based retrieval agents.
\end{itemize}

\section{Related Work}

\subsection{Supervised Fine-Tuning}

Supervised fine-tuning (SFT) remains a central paradigm for adapting large language models to instruction following, structured output generation, and complex multi-step reasoning. Recent studies show that supervising models with explicit reasoning trajectories or intermediate decision states significantly improves reasoning stability and error localization, particularly in long-horizon tasks\cite{Du2024TrajAgentAL,Sullivan2025ProceduralEG,Yu2025LongShortCM,DBLP:journals/corr/abs-2604-03647,xu2026visualsimilarityentityalignedretrieval}.

Beyond static reasoning traces, recent work has explored trajectory-style supervision for learning tool use and environment interaction. Agent-oriented SFT frameworks show that models can acquire structured reasoning–action patterns when trained on annotated interaction trajectories, enabling effective tool invocation, planning, and feedback integration\cite{Chen2025StepwiseAI,Liu2025UFTUS,wu2026promsa,wu2026models}. These results indicate that modern SFT, when combined with trajectory-level supervision, provides a strong foundation for learning structured reasoning and tool-augmented problem solving.

\subsection{Search Agents}

Recent work has enabled search-based agents that call external tools, issue search queries during reasoning, and iteratively incorporate retrieved evidence for multi-step inference\cite{He2025PaSaAL,Li2025ReSeekAS,Wu2025PORToolTL,DBLP:conf/aaai/WuZJXLZLJW26,DBLP:journals/corr/abs-2602-00104}. 
Compared with conventional one-shot RAG, this line of work emphasizes adaptive tool use, sequential evidence acquisition, and trajectory-level decision making.

Most existing search-agent frameworks focus on open-world information seeking, where the main challenge is to formulate effective queries and gather useful evidence from noisy environments\cite{Luo2025BrowsingLH,Li2024BenchmarkingMR,Zheng2025DeepResearcherSD}. 
Recent multimodal agent and planning work further studies multi-step tool use, but is generally evaluated in open-world or general-purpose settings where the target entities are often common or explicitly specified\cite{Wang2024MLLMToolAM}.

In contrast, KB-VQA often starts from fine-grained or long-tail visual entities that are not explicitly named in the question. 
DBAgent therefore focuses on KB-VQA-specific decisions over the current multimodal state: whether to answer directly, use image retrieval for entity grounding, use text retrieval for factual lookup, or stop searching.

\subsection{KB-VQA Retrieval-Augmented Methods}

KB-VQA targets questions whose answers depend on long-tail, fine-grained encyclopedic knowledge that is not directly observable in the image. 
Recent studies therefore treat KB-VQA as a multimodal RAG problem\cite{Zhang2025FineGrainedKS,Su2024SKVQASK}: the system retrieves candidate evidence (typically from large-scale resources such as Wikipedia articles and associated images) and conditions the VLM on the retrieved content to improve coverage on underrepresented queries.
Several works improve KB-VQA by enhancing multimodal retrieval or post-retrieval filtering.
Nevertheless, they still rely on a static workflow, where retrieval is fixed and independent of the model’s reasoning process\cite{Yan2024EchoSightAV,Cocchi2024AugmentingML}. Retrieval is usually performed in a predefined manner, followed by filtering and answer generation. This design assumes all questions follow the same reasoning process, which is often not true. 
This motivates us to move beyond fixed pipelines and propose an agent-based formulation of KB-VQA.

\section{Methods}

We propose a search-agent framework for knowledge-intensive KB-VQA. 
The core idea is to reformulate KB-VQA from a fixed \textit{retrieve--then-generate} pipeline into a multi-step decision-making process. 
Instead of following a predefined workflow, the model dynamically assesses whether the current information state is sufficient for answering, determines which tool to invoke, formulates appropriate queries, and decides when to stop searching.
As a result, the search depth adapts to the complexity of questions, avoiding unnecessary tool calls and premature termination.
We first formalize the problem in Section~\ref{sec:formulation}, then describe the inference-time execution process and action semantics in Section~\ref{sec:rollout}, and finally present the trajectory construction procedure and supervised fine-tuning objective in Section~\ref{sec:training}.
% --------------------------------------------------------

\subsection{Problem Formulation}
\label{sec:formulation}

Given an input image $I$ and a natural language question $q$, the goal of KB-VQA is to produce the answer $y$.
Instead of modeling this process as a single-step mapping, we formulate it as a sequential decision-making problem.
The model interacts with an external knowledge environment over multiple steps, gradually collecting evidence and refining its understanding.
At step $t$, the model maintains a state $s_t$, which represents all the information it has observed so far:
\begin{equation}
s_t \triangleq (I, q, a_1, e_1, \ldots, a_{t-1}, e_{t-1}),
\end{equation}
where $a_i$ denotes the action and $e_i$ denotes the evidence returned by tool calls in  the previous step.
Based on the current state $s_t$, the model selects an action $a_t$ from a predefined action space $\mathcal{A}$.
If the selected action involves an external tool, the environment returns new evidence $e_t$, which is appended to the state:
\begin{equation}
s_{t+1} = (s_t, a_t, e_t).
\end{equation}
This process continues until the model outputs an answer or reaches a maximum step budget $B$.
The resulting sequence forms a complete search trajectory:
\begin{equation}
\tau = (I, q, a_1, e_1, a_2, e_2, \ldots, a_T),
\end{equation}

% --------------------------------------------------------

\subsection{Dynamic Search-Agent Framework}
\label{sec:rollout}

\paragraph{Overall Framework.}
As shown in Figure~\ref{fig:pipeline}, our inference follows an iterative \textit{reason--act--observe--update} loop. The detailed inference procedure is summarized in Algorithm~\ref{alg:rollout} in Appendix~\ref{sec:appendix}.
The reasoning process is explicitly generated between the \tagReason{<think>} and \tagReason{</think>} tags.
Our framework allows the model to dynamically decide:
(1) whether the current state is sufficient for answering,
(2) what type of information to acquire and which tool to use, and
(3) when to stop and generate the final answer.
This adaptive mechanism enables the model to adjust its search depth based on question difficulty, producing reasoning trajectories with matched complexity.

\paragraph{Action Selection.}
Based on its reasoning, the model selects one action from a predefined action space.
Importantly, this action space is designed to reflect the intrinsic challenges of KB-VQA. 
For example, in many cases, the question does not explicitly mention the target entity, and the model must first infer the entity from the image before it can issue a reliable textual query. 
As a result, the actions are not independent, but form a collaborative toolset. 

Formally, the action space is defined as:
\begin{equation}
\mathcal{A} = \left\{ a^{\text{ans}}, a^{\text{text}}, a^{\text{img}}, a^{\text{cap}} \right\}.
\end{equation}
Each action corresponds to a specific form of interaction with the external knowledge environment:
\begin{itemize}
  \item \textbf{Answer} (\(a^{\text{ans}}\)):  
  If the search agent determines that the accumulated evidence is sufficient, it outputs the final answer enclosed within the \tagAnswer{<answer>} and \tagAnswer{</answer>} tags.

  \item \textbf{Text Search} (\(a^{\text{text}}\)):  
  The search agent generates a natural-language query, enclosed within the \tagSearch{<text\_search>} and \tagSearch{</text\_search>} tags, to retrieve relevant textual passages from the knowledge base.

  \item \textbf{Image Search} (\(a^{\text{img}}\)):  
  The search agent queries the visual retrieval module using the \tagSearch{<image\_search>} and \tagSearch{</image\_search>} tags, which returns visually similar images together with their associated textual descriptions.

  \item \textbf{Caption} (\(a^{\text{cap}}\)):  
  When the current multimodal representation is insufficient to form a reliable retrieval query---for example, when the target entity cannot be confidently identified---the model may generate a task-oriented caption enclosed within the \tagCap{<caption>} and \tagCap{</caption>} tags. 
  This caption abstracts the visual content into a textual form and serves as an intermediate semantic representation for subsequent query rewriting and retrieval.
\end{itemize}

% --------------------------------------------------------

\paragraph{State Update.}
When a tool action is selected, the retrieved results are wrapped into structured evidence and enclosed within the \tagEv{<evidence>} and \tagEv{</evidence>} tags.
This evidence is then appended to the state to form $s_{t+1}$.
The caption action does not introduce new external evidence, but it is appended to the internal context as a generated intermediate representation. This design keeps the semantics of state transitions clear: only interactions with the environment update the external evidence buffer, while captioning refines the internal textual state.

% --------------------------------------------------------

\subsection{Trajectory Construction and Training}
\label{sec:training}

Although modern multimodal models are strong at multi-step reasoning, instruction following, and cross-modal generation, these abilities are usually learned as isolated skills rather than as a unified decision policy.
In KB-VQA, the model must not only answer questions, but also repeatedly assess its information state and decide whether to retrieve, how to retrieve, and when to stop.
We observe that without explicit training on this decision process, models struggle to coordinate these abilities into stable and controllable behaviors.

\subsubsection{Failure-Aware Trajectory Branching}

We design a failure-aware branching mechanism based on common failure patterns in KB-VQA.
For each original image--question--answer sample, we first test whether the model can answer correctly without using any external tools.
Such samples are treated as \textit{directly answerable} and used to construct zero-retrieval trajectories.
Gold answers and entity annotations are used only offline to judge training trajectories and route failed attempts, and are never available during inference.

If the model fails, indicating that the current information state is insufficient, we further categorize the failure into two types:
(1) the entity is recognized but the required factual or reasoning support is missing, and
(2) the entity cannot be reliably identified.
In the first case, the model has a plausible visual grounding but lacks sufficient support to answer the question, so text retrieval is used to access relevant external information.
In the second case, the model lacks a clear semantic anchor, so visual retrieval and a \tagCap{<caption>} step are used to support subsequent query reformulation.

% --------------------------------------------------------

\subsubsection{Difficulty Modeling}

Questions in KB-VQA differ significantly in search depth, tool coordination, and reasoning structure.
This forms a distribution with varying levels of \textbf{decision complexity} across different problems.
To capture this structure, we partition the dataset into multiple difficulty subsets:
\begin{equation}
\mathcal{D} = \bigcup_{k=1}^{K} \mathcal{D}_k.
\end{equation}

We characterize the difficulty of KB-VQA problems along three key dimensions:
(1) search depth,
(2) tool usage structure, and
(3) intermediate reasoning patterns.
Based on these factors, we group trajectories into easy, medium, and hard subsets.
From each subset, we sample a fixed number of trajectories for training.

% --------------------------------------------------------

\subsubsection{Training Template and SFT Objective}

To ensure the model learns decision behaviors rather than memorizing retrieved content, we use a unified instruction template to regulate reasoning, action selection, and output formats in a structured and consistent manner.
At each step, the model must choose exactly one action and explicitly state its reasoning between the \tagReason{<think>} and \tagReason{</think>} tags.

If the current information is sufficient, the model outputs the final answer within \tagAnswer{<answer>} and \tagAnswer{</answer>}.
Otherwise, it selects \tagSearch{<text\_search>}, \tagSearch{<image\_search>} or \tagCap{<caption>}.
The \tagCap{<caption>} action serves as an intermediate representation for subsequent query rewriting and retrieval.

During training, each trajectory is linearized into a token sequence:
\begin{equation}
(u, a_1, o_1, \ldots, a_n),
\end{equation}
where $u$ is the initial instruction, $a_i$ is the action token, and $o_i$ is the environment observation enclosed within the \tagEv{<evidence>} and \tagEv{</evidence>} tags for each step.
We mask all observation tokens and apply supervision only to decision-related tokens.
The loss is defined as:
\begin{equation}
\mathcal{L} = -\sum_j \log p_\theta(t_j \mid t_{<j}) \cdot \mathbb{I}(t_j \in \mathcal{Y}),
\end{equation}
where $\mathcal{Y}$ denotes the set of decision tokens.

This shifts supervision from final-answer prediction to learning the sequence of decisions that lead to the answer.
Although trained via SFT, DBAgent learns a decision policy rather than following a fixed pipeline, allowing it to adapt its retrieval strategy to unseen decision states during inference.

\begin{table*}[t]
\centering
\small
\setlength{\tabcolsep}{3.8pt}
\renewcommand{\arraystretch}{1.12}

\resizebox{\textwidth}{!}{%

\begin{tabular}{l l l c c c c c}
\toprule
\multirow{2}{*}{Method} & \multirow{2}{*}{LLM} & \multirow{2}{*}{Retriever}
& \multicolumn{2}{c}{E-VQA} & \multicolumn{3}{c}{InfoSeek} \\
\cmidrule(lr){4-5} \cmidrule(lr){6-8}
& & & Single-Hop & All & Unseen-Q & Unseen-E & All \\
\midrule

% ---------- Zero-shot ----------
\rowcolor{lightgray}
\multicolumn{8}{l}{\textbf{Zero-shot MLLMs}} \\
BLIP-2 & Flan-T5$_{XL}$ & -- & 12.6 & 12.4 & 12.7 & 12.3 & 12.5 \\
InstructBLIP & Flan-T5$_{XL}$ & -- & 11.9 & 12.0 & 8.9 & 7.4 & 8.1 \\
LLaVA-v1.5 & Vicuna-7B & -- & 16.3 & 16.9 & 9.6 & 9.4 & 9.5 \\
GPT-4V & GPT-4V & -- & 26.9 & 28.1 & 15.0 & 14.3 & 14.6 \\
Qwen2.5-VL-7B (Base) & Qwen2.5-VL-7B (Base) & -- & 21.7 & 20.3 & 22.8 & 24.1 & 23.7 \\

\midrule
% ---------- Classical RAG ----------
\rowcolor{lightgray}
\multicolumn{8}{l}{\textbf{Classical Retrieval-Augmented Models}} \\
DPR & Multi-passage BERT & CLIP ViT-B/32 & 29.1 & -- & -- & -- & 12.4 \\
RORA-VLM & LLaVA-v1.5-7B & CLIP ViT-L/14 & -- & 20.3 & 25.1 & 27.3 & -- \\
EchoSight & Mistral-7B/LLaMA-3-8B & EVA-CLIP-8B & 19.4 & -- & -- & -- & 27.7 \\
Wiki-LLaVA & LLaVA-v1.5-7B & CLIP ViT-L/14 + Contriever & 17.7 & 20.3 & 30.1 & 27.8 & 28.9 \\

\midrule
% ---------- Reasoning / RL ----------
\rowcolor{lightgray}
\multicolumn{8}{l}{\textbf{Retrieval-Augmented Models with Reasoning / RL}} \\
ReflectiVA & LLaMA-3.1-8B & EVA-CLIP-8B & 28.0 & 29.2 & 40.4 & 39.8 & 40.1 \\
ReflectiVA & Qwen2.5-VL-7B & EVA-CLIP-8B & 36.8 & 36.8 & 43.5 & 44.3 & 43.9 \\
VLM-PRF & Qwen2.5-VL-7B & EVA-CLIP-8B & 37.1 & 36.0 & 43.3 & 42.7 & 42.8 \\
VLM-PRF & InternVL3-8B & EVA-CLIP-8B & 40.1 & 39.2 & 43.5 & 42.1 & 42.5 \\

\midrule
% ---------- Ours ----------
\rowcolor{lightgray}
\multicolumn{8}{l}{\textbf{Ours: Agent-based Search}} \\
\rowcolor{lightblue}
DBAgent (w/o SFT) & Qwen2.5-VL-7B & EVA-CLIP-8B + BGE-M3 & 23.6 & 23.9 & 24.9 & 24.2 & 24.4 \\
\rowcolor{lightblue}
DBAgent (SFT: InfoSeek) & Qwen2.5-VL-7B & EVA-CLIP-8B + BGE-M3 & 45.1 & 44.3 & \textbf{46.5} & \textbf{51.0} & \textbf{49.9} \\
\rowcolor{lightblue}
DBAgent (SFT: E-VQA) & Qwen2.5-VL-7B & EVA-CLIP-8B + BGE-M3 & \textbf{46.5} & \textbf{45.8} & 43.1 & 50.2 & 48.4 \\
\rowcolor{lightblue}
DBAgent (SFT: Mixed) & Qwen2.5-VL-7B & EVA-CLIP-8B + BGE-M3 & 46.0 & 45.2 & 43.6 & 50.4 & 48.7 \\

\bottomrule
\end{tabular}

}

\caption{Main results on E-VQA and InfoSeek. We report VQA accuracy (\%) under different LLMs and retrievers. Our agent-based methods are highlighted in light blue.}
\label{tab:main_results}
\end{table*}

\section{Experiment}
\subsection{Experimental Setup}

\paragraph{Datasets.}
We evaluate our method on two knowledge-intensive visual question answering benchmarks: InfoSeek~\cite{Chen2023CanPV} and Encyclopedic-VQA (E-VQA)~\cite{Mensink2023EncyclopedicVV}. 
Both datasets require models to answer image-based questions by leveraging external encyclopedic knowledge.
InfoSeek consists of approximately 1.3M image--question--answer triplets associated with around 11K Wikipedia pages. 
Following prior work, the validation set is categorized into two subsets, namely \textit{Unseen-Entity} and \textit{Unseen-Question}, which evaluate the model’s generalization ability to novel entities and novel question formulations in practice.
Encyclopedic-VQA (E-VQA) contains over 221K question--answer pairs, each linked to up to five images and covering approximately 16.7K fine-grained entities corresponding to Wikipedia pages. 
Following standard practice, we report results on the official test set.

\paragraph{Baselines.}
We compare DBAgent with diverse baselines covering different modeling paradigms in our experiments. 
These include multimodal large language models that directly answer questions, retrieval-augmented generation models, and models that incorporate explicit reasoning mechanisms.

\paragraph{Evaluation Metrics.}
We evaluate both answer quality and retrieval quality, following the official protocols of each dataset.
For InfoSeek, we use Exact Match (EM) as the primary metric. A prediction is considered correct if it exactly matches any of the ground-truth answers. 
For E-VQA, we adopt the BERT-based Matching (BEM) score\cite{Bulian2022TomaytoTB}, which measures the semantic similarity between predicted answers and ground-truth answers.
To assess retrieval performance, we report a hit-based retrieval accuracy, which measures whether the ground-truth Wikipedia article is successfully covered by the retrieved results at any turn. This metric reflects how effectively the retrieval module provides relevant external evidence for downstream reasoning and answer generation.

\subsection{Implementation Details}

\paragraph{Knowledge Base and Retrieval Tools.}
We construct an external knowledge base from Wikipedia and support both image-level and text-level retrieval. For image retrieval, we follow an EchoSight-style indexing strategy by encoding all Wikipedia images into a dense vector space and maintaining a mapping from each image to its source article\cite{Yan2024EchoSightAV}. Given a query image, the system retrieves the most similar results and returns the associated Wikipedia article as candidate evidence. In all experiments, we use the top-$k$ results with $k=1$.
For text retrieval, we segment each Wikipedia article into multiple textual sections and encode all sections using BGE embeddings. Retrieval is performed at the section level, and the top-$k$ most relevant sections are returned as textual evidence, with $k=3$. 
We use Qwen2.5-VL-7B-Instruct as the backbone model for both training and inference in all experiments.

\paragraph{Training Setup.}
We fine-tune Qwen2.5-VL-7B using supervised fine-tuning on our trajectory-based training datasets. The trajectories are grouped into three difficulty levels, namely easy, medium, and hard, according to question complexity and required retrieval depth, and are sampled with a balanced ratio of 1:1:1. They are constructed from the InfoSeek and E-VQA, yielding approximately 160K samples from each dataset.
To prevent the model from memorizing retrieved content, we apply a loss-masking strategy that excludes tool-returned evidence tokens from supervision. 
Additional hyperparameter choices and implementation details are provided in the supplementary material.
All data construction procedures are conducted exclusively on the standard training split and do not involve any information from the evaluation sets.

\begin{table}[t]
\centering
\small
\setlength{\tabcolsep}{8pt}
\renewcommand{\arraystretch}{1.15}

\resizebox{\columnwidth}{!}{%
\begin{tabular}{c c c c}
\toprule
Trajectory Type & Sample Proportion & Retrieval Recall & Answer Accuracy \\
\midrule
A                     & 5.4  & --   & 69.7 \\
I$\rightarrow$A       & 25.7 & 65.9 & 56.0 \\
T$\rightarrow$A       & 36.1 & 81.6 & 49.5 \\
I$\rightarrow$T$\rightarrow$A & 15.7 & 55.2 & 43.5 \\
T$\rightarrow$T$\rightarrow$A & 17.1 & 70.3 & 41.1 \\
\bottomrule
\end{tabular}%
}

\caption{Trajectory-level analysis on the test set.}
\label{tab:trajectory_analysis}
\end{table}

\subsection{Main Results and Analyses}

\paragraph{Overall Results on Encyclopedic-VQA and InfoSeek}

We evaluate our method on two representative KB-VQA benchmarks in the experiments.
The results are shown in Table \ref{tab:main_results}. 
We train our models DBAgent under three data settings: InfoSeek, E-VQA, and a mixture of the two.
Across the evaluated settings, DBAgent consistently outperforms prior baselines, while different SFT data settings show different strengths. Specifically, DBAgent with E-VQA SFT reaches 45.8\% on E-VQA (All), outperforming the strongest baseline by about 6 points. On InfoSeek (All), DBAgent with InfoSeek SFT achieves 49.9\%, which is also substantially higher than existing methods. The mixed model provides a strong cross-dataset trade-off, achieving 45.2\% on E-VQA (All) and 48.7\% on InfoSeek (All). Moreover, DBAgent maintains clear advantages on the Unseen-Q and Unseen-E splits of InfoSeek.
These results indicate that the DBAgent not only improves overall answer accuracy, but also generalizes better to unseen question forms and previously unseen entities.

\begin{table}[t]
\centering
\small
\setlength{\tabcolsep}{4.5pt}
\renewcommand{\arraystretch}{1.12}

\resizebox{\columnwidth}{!}{%
\begin{tabular}{l l c ccc}
\toprule
\multirow{2}{*}{Model} & \multirow{2}{*}{Generator}
& \multicolumn{1}{c}{E-VQA} & \multicolumn{3}{c}{InfoSeek} \\
\cmidrule(lr){3-3} \cmidrule(lr){4-6}
& & Single-Hop & Un-Q & Un-E & All \\
\midrule

% -------- Baselines --------
Qwen2.5-VL-3B~ & Qwen2.5-3B  & 72.1 & 47.0 & 43.0 & 44.9 \\
Qwen2.5-VL-7B~ & Qwen2.5-7B  & 78.3 & 41.6 & 41.3 & 41.4 \\
ReflectiVA~    & Qwen2.5-VL-3B & 72.9 & 53.4 & 53.9 & 53.7 \\
\midrule
Wiki-LLaVA~   & LLaVA-v1.5-7B & 38.5 & 52.7 & 50.3 & 51.5 \\
ReflectiVA~   & LLaVA-MORE-8B & 75.2 & 57.8 & 57.4 & 57.6 \\
ReflectiVA~   & Qwen2.5-VL-7B & 71.3 & 56.0 & 56.0 & 56.0 \\
\textbf{ReAG}~& Qwen2.5-VL-7B & 81.5 & 60.7 & 58.9 & 59.7 \\
\midrule

% -------- Our variants --------
\rowcolor{lightblue}
\textbf{DBAgent (SFT: InfoSeek)} & Qwen2.5-VL-7B & 80.2 & 60.0 & 61.4 & 61.1 \\
\rowcolor{lightblue}
\textbf{DBAgent (SFT: E-VQA)} & Qwen2.5-VL-7B & 79.4 & 59.6 & 60.0 & 59.9 \\
\rowcolor{lightblue}
\textbf{DBAgent (SFT: Mixed)} & Qwen2.5-VL-7B & 80.9 & 60.2 & 62.7 & 62.1 \\

\end{tabular}%
}

\caption{VQA accuracy scores on Encyclopedic-VQA and InfoSeek with oracle Wikipedia pages.}
\label{tab:singlecol_comp}
\end{table}

\begin{table}[t]
\centering
\small
\setlength{\tabcolsep}{6pt}
\renewcommand{\arraystretch}{1.15}
\resizebox{\columnwidth}{!}{%
\begin{tabular}{l c c}
\toprule
Method & E-VQA (All) & InfoSeek (All) \\
\midrule
No Retrieval & 20.3 & 23.7 \\
Forced Image Retrieval & 22.5 & 27.7 \\
Caption-based RAG & 24.7 & 30.5 \\
\midrule
\textbf{DBAgent (SFT: Mixed)} & 45.2 & 48.7 \\
\bottomrule
\end{tabular}
}
\caption{Ablation study on different retrieval and decision strategies. We report the overall accuracy on E-VQA and InfoSeek.}
\vspace{-2mm}
\label{tab:ablation}
\end{table}

\paragraph{Trajectory Distribution and Difficulty Stratification}

We analyze the distribution of different reasoning trajectories, as shown in Table \ref{tab:trajectory_analysis}. 
Here, I, T, and A denote image retrieval, text retrieval, and answer generation, respectively.
As the trajectory becomes longer, both retrieval recall and answer accuracy tend to decrease. This is because longer trajectories usually correspond to more difficult questions: although additional retrieval steps are expected to provide more evidence, these samples often involve harder entity grounding, noisier retrieval, or more complex reasoning. 
These results indicate that a fixed-depth retrieval strategy is not suitable for KB-VQA. Instead, adaptively adjusting the search depth better matches the task structure.

\paragraph{Oracle Analysis with Ground-Truth Wikipedia Pages}

Under the oracle setting, where the ground-truth Wikipedia pages are directly provided, DBAgent still maintains a clear advantage as shown in Table \ref{tab:singlecol_comp}.
This suggests that our gains are not solely due to improved retrieval, but also come from more effective evidence selection and evidence-grounded reasoning.

\subsection{Ablation Studies and Behavioral Analysis}

\paragraph{Ablation on Retrieval and Decision Strategies}

As shown in Table~\ref{tab:ablation}, simply adding retrieval does not always lead to better performance.
In particular, two fixed retrieval variants show only marginal gains: (i) performing image retrieval for every sample and using the Top-1 matched document, and (ii) first generating a caption and then conducting text retrieval with Top-3 passages. These results suggest that DBAgent’s advantage lies in deciding when and how to retrieve evidence.

\paragraph{Impact of Knowledge Base Size}

\begin{figure}[t]
    \centering
    \includegraphics[width=\linewidth]{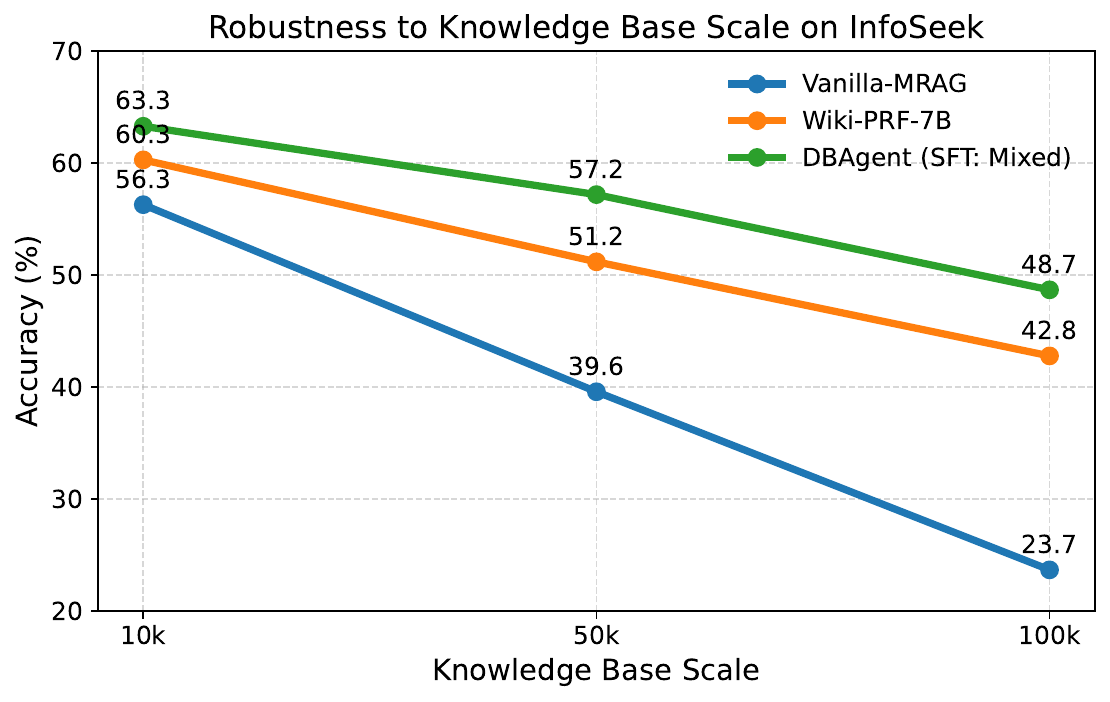}
    \caption{Ablation studies on the scale of knowledge base on InfoSeek.}
    \label{fig:kb_scale}
\end{figure}

We further analyze the impact of the knowledge base size. As shown in Figure~\ref{fig:kb_scale}, when the size increases from 10k to 100k, the performance of all methods drops.
In contrast, DBAgent shows a much smoother degradation, dropping from 63.3\% to 48.7\%. This indicates that our method is more robust in high-noise scenarios, further supporting that decision-based search is more suitable for KB-VQA.

\paragraph{Impact of Different Top-$k$ Settings}

As shown in Table~\ref{tab:topk_ablation_side_by_side}, Top-$k$ selection involves a trade-off between performance and cost.
A smaller Top-$k$ may limit retrieval recall, while a larger Top-$k$ increases context length and inference cost.
Since increasing the number of retrieved text passages from 3 to 5 brings only marginal gains, we choose text Top-$k=3$ and image Top-$k=1$ as the final setting for a better balance between accuracy and efficiency.
Overall, DBAgent remains stable across different Top-$k$ combinations.

\begin{table}[t]
\centering
\small
\setlength{\tabcolsep}{6pt}
\renewcommand{\arraystretch}{1.15}

\begin{minipage}[t]{0.49\linewidth}
\centering
\begin{tabular}{c c c c}
\toprule
\multirow{2}{*}{Text $k$} & \multicolumn{3}{c}{Image $k$} \\
\cmidrule(lr){2-4}
 & 1 & 2 & 3 \\
\midrule
1 & 40.4 & 40.2 & 39.8 \\
3 & 45.2 & 45.1 & 43.7 \\
5 & 45.2 & 44.9 & 40.6 \\
\bottomrule
\end{tabular}

\vspace{2pt}
{\footnotesize (a) E-VQA\par}
\end{minipage}\hfill
\begin{minipage}[t]{0.49\linewidth}
\centering
\begin{tabular}{c c c c}
\toprule
\multirow{2}{*}{Text $k$} & \multicolumn{3}{c}{Image $k$} \\
\cmidrule(lr){2-4}
 & 1 & 2 & 3 \\
\midrule
1 & 44.4 & 43.9 & 42.6 \\
3 & 48.7 & 48.2 & 46.6 \\
5 & 49.1 & 47.5 & 46.3 \\
\bottomrule
\end{tabular}

\vspace{2pt}
{\footnotesize (b) InfoSeek\par}
\end{minipage}

\caption{Impact of retrieval top-$k$ for text and image retrieval. We report overall accuracy (\%) on E-VQA and InfoSeek.}
\vspace{-2mm}
\label{tab:topk_ablation_side_by_side}
\end{table}

\section{Conclusion}
We present DBAgent, a decision-based search agent framework for knowledge-based visual question answering (KB-VQA). Instead of following a fixed retrieval-then-generation pipeline, DBAgent formulates KB-VQA as a multi-step decision-making process, where the model dynamically selects among different actions based on its current information state.
To enable effective learning of such behaviors, we further propose an automated trajectory construction framework and build a large-scale, high-quality training dataset that explicitly records reasoning steps, tool usage, and intermediate decisions. This dataset provides structured supervision for training decision-aware agents.
Extensive experiments on InfoSeek and E-VQA demonstrate that DBAgent consistently outperforms existing retrieval-augmented baselines, confirming the effectiveness and feasibility of our framework.

\section{Limitations}

\paragraph{Trajectory Supervision}
DBAgent is trained with automatically constructed reasoning--retrieval trajectories, which enables scalable supervision without manual trajectory annotation. While our trajectory-quality analysis suggests that the constructed trajectories provide useful evidence-grounded supervision, future work could further incorporate human verification or finer-grained trajectory quality control.

\paragraph{Inference Cost}
DBAgent performs iterative reasoning and retrieval during inference, which introduces additional cost compared with single-pass methods. Our efficiency analysis shows that the overhead is moderate under the predefined tool budget, and future work may further improve efficiency through adaptive stopping or more selective tool use.

\section{Acknowledgements}

This work is supported by the NSFC fund (62576190), in part by the Shenzhen Science and Technology Project under Grant (KJZD20240903103210014)


\bibliography{custom}

@article{Tanaka2025VDocRAGRG,
  title={VDocRAG: Retrieval-Augmented Generation over Visually-Rich Documents},
  author={Ryota Tanaka and Taichi Iki and Taku Hasegawa and Kyosuke Nishida and Kuniko Saito and Jun Suzuki},
  journal={2025 IEEE/CVF Conference on Computer Vision and Pattern Recognition (CVPR)},
  year={2025},
  pages={24827-24837},
  url={https://api.semanticscholar.org/CorpusID:277781495}
}

@article{Xiao2024VideoQAIT,
  title={VideoQA in the Era of LLMs: An Empirical Study},
  author={Junbin Xiao and Nanxin Huang and Hangyu Qin and Dongyang Li and Yicong Li and Fengbin Zhu and Zhulin Tao and Jianxing Yu and Liang Lin and Tat-Seng Chua and Angela Yao},
  journal={International Journal of Computer Vision},
  year={2024},
  volume={133},
  pages={3970 - 3993},
  url={https://api.semanticscholar.org/CorpusID:271768879}
}

@article{Long2025RetrievalAugmentedVQ,
  title={Retrieval-Augmented Visual Question Answering via Built-in Autoregressive Search Engines},
  author={Xinwei Long and Zhiyuan Ma and Ermo Hua and Kaiyan Zhang and Biqing Qi and Bowen Zhou},
  journal={ArXiv},
  year={2025},
  volume={abs/2502.16641},
  url={https://api.semanticscholar.org/CorpusID:276576103}
}

@article{Choi2025MultimodalIR,
  title={Multimodal Iterative RAG for Knowledge-Intensive Visual Question Answering},
  author={Changin Choi and Wonseok Lee and Jungmin Ko and Wonjong Rhee},
  journal={ArXiv},
  year={2025},
  volume={abs/2509.00798},
  url={https://api.semanticscholar.org/CorpusID:281080708}
}

@article{Hong2025KnowledgebasedVQ,
  title={Knowledge-based Visual Question Answer with Multimodal Processing, Retrieval and Filtering},
  author={Yuyang Hong and Jiaqi Gu and Qi Yang and Lubin Fan and Yue Wu and Ying Wang and Kun Ding and Shiming Xiang and Jieping Ye},
  journal={ArXiv},
  year={2025},
  volume={abs/2510.14605},
  url={https://api.semanticscholar.org/CorpusID:282138886}
}

@inproceedings{Du2024TrajAgentAL,
  title={TrajAgent: An LLM-Agent Framework for Trajectory Modeling via Large-and-Small Model Collaboration},
  author={Yuwei Du and Jie Feng and Jie Zhao and Yong Li},
  year={2024},
  url={https://api.semanticscholar.org/CorpusID:279075751}
}

@article{Sullivan2025ProceduralEG,
  title={Procedural Environment Generation for Tool-Use Agents},
  author={Michael Sullivan and Mareike Hartmann and Alexander Koller},
  journal={ArXiv},
  year={2025},
  volume={abs/2506.11045},
  url={https://api.semanticscholar.org/CorpusID:279392238}
}

@article{Yu2025LongShortCM,
  title={Long-Short Chain-of-Thought Mixture Supervised Fine-Tuning Eliciting Efficient Reasoning in Large Language Models},
  author={Bin Yu and Hang Yuan and Haotian Li and Xueyin Xu and Yuliang Wei and Bailing Wang and Weizhen Qi and Kai Chen},
  journal={ArXiv},
  year={2025},
  volume={abs/2505.03469},
  url={https://api.semanticscholar.org/CorpusID:278339126}
}

@article{Liu2025UFTUS,
  title={UFT: Unifying Supervised and Reinforcement Fine-Tuning},
  author={Mingyang Liu and Gabriele Farina and Asuman E. Ozdaglar},
  journal={ArXiv},
  year={2025},
  volume={abs/2505.16984},
  url={https://api.semanticscholar.org/CorpusID:278789575}
}

@article{Chen2025StepwiseAI,
  title={Step-wise Adaptive Integration of Supervised Fine-tuning and Reinforcement Learning for Task-Specific LLMs},
  author={Jack Xiaoyu Chen and Fazhong Liu and Na Liu and Yu Luo and Erqu Qin and Harry Zheng and Tian Dong and Haojin Zhu and Yan Meng and Xiao Wang},
  journal={ArXiv},
  year={2025},
  volume={abs/2505.13026},
  url={https://api.semanticscholar.org/CorpusID:278740779}
}

@inproceedings{He2025PaSaAL,
  title={PaSa: An LLM Agent for Comprehensive Academic Paper Search},
  author={Yichen He and Guanhua Huang and Peiyuan Feng and Yuan Lin and Yuchen Zhang and Hang Li and Weinan E},
  booktitle={Annual Meeting of the Association for Computational Linguistics},
  year={2025},
  url={https://api.semanticscholar.org/CorpusID:275606590}
}

@article{Li2025ReSeekAS,
  title={ReSeek: A Self-Correcting Framework for Search Agents with Instructive Rewards},
  author={Shiyu Li and Yang Tang and Yifan Wang and Peiming Li and Xi Chen},
  journal={ArXiv},
  year={2025},
  volume={abs/2510.00568},
  url={https://api.semanticscholar.org/CorpusID:281705848}
}

@article{Wu2025PORToolTL,
  title={PORTool: Tool-Use LLM Training with Rewarded Tree},
  author={Feijie Wu and Wei Zhu and Yuxiang Zhang and Soumya Chatterjee and Jiarong Zhu and Fan Mo and Rodin Luo and Jing Gao},
  journal={ArXiv},
  year={2025},
  volume={abs/2510.26020},
  url={https://api.semanticscholar.org/CorpusID:282592220}
}

@inproceedings{Luo2025BrowsingLH,
  title={Browsing Like Human: A Multimodal Web Agent with Experiential Fast-and-Slow Thinking},
  author={Haohao Luo and Jiayi Kuang and Wei Liu and Ying Shen and Jian Luan and Yang Deng},
  booktitle={Annual Meeting of the Association for Computational Linguistics},
  year={2025},
  url={https://api.semanticscholar.org/CorpusID:280017117}
}

@article{Li2024BenchmarkingMR,
  title={Benchmarking Multimodal Retrieval Augmented Generation with Dynamic VQA Dataset and Self-adaptive Planning Agent},
  author={Yangning Li and Yinghui Li and Xinyu Wang and Yong Jiang and Zhen Zhang and Xinran Zheng and Hui Wang and Hai-Tao Zheng and Pengjun Xie and Philip S. Yu and Fei Huang and Jingren Zhou},
  journal={ArXiv},
  year={2024},
  volume={abs/2411.02937},
  url={https://api.semanticscholar.org/CorpusID:273821256}
}

@inproceedings{Zheng2025DeepResearcherSD,
  title={DeepResearcher: Scaling Deep Research via Reinforcement Learning in Real-world Environments},
  author={Yuxiang Zheng and Dayuan Fu and Xiangkun Hu and Xiaojie Cai and Lyumanshan Ye and Pengrui Lu and Pengfei Liu},
  booktitle={Conference on Empirical Methods in Natural Language Processing},
  year={2025},
  url={https://api.semanticscholar.org/CorpusID:277596185}
}

@article{Wang2024MLLMToolAM,
  title={MLLM-Tool: A Multimodal Large Language Model for Tool Agent Learning},
  author={Chenyu Wang and Weixin Luo and Qianyu Chen and Haonan Mai and Jindi Guo and Sixun Dong and Xi Xuan and Zhengxin Li and Lin Ma and Shenghua Gao},
  journal={2025 IEEE/CVF Winter Conference on Applications of Computer Vision (WACV)},
  year={2024},
  pages={6678-6687},
  url={https://api.semanticscholar.org/CorpusID:267060838}
}

@inproceedings{Zhang2025FineGrainedKS,
  title={Fine-Grained Knowledge Structuring and Retrieval for Visual Question Answering},
  author={Zhengxuan Zhang and Yin Wu and Yuyu Luo and Nan Tang},
  year={2025},
  url={https://api.semanticscholar.org/CorpusID:276724839}
}

@inproceedings{Yan2024EchoSightAV,
  title={EchoSight: Advancing Visual-Language Models with Wiki Knowledge},
  author={Yibin Yan and Weidi Xie},
  booktitle={Conference on Empirical Methods in Natural Language Processing},
  year={2024},
  url={https://api.semanticscholar.org/CorpusID:271244745}
}

@article{Cocchi2024AugmentingML,
  title={Augmenting Multimodal LLMs with Self-Reflective Tokens for Knowledge-based Visual Question Answering},
  author={Federico Cocchi and Nicholas Moratelli and Marcella Cornia and Lorenzo Baraldi and Rita Cucchiara},
  journal={2025 IEEE/CVF Conference on Computer Vision and Pattern Recognition (CVPR)},
  year={2024},
  pages={9199-9209},
  url={https://api.semanticscholar.org/CorpusID:274280457}
}

@article{Chen2023CanPV,
  title={Can Pre-trained Vision and Language Models Answer Visual Information-Seeking Questions?},
  author={Yang Chen and Hexiang Hu and Yi Luan and Haitian Sun and Soravit Changpinyo and Alan Ritter and Ming-Wei Chang},
  journal={ArXiv},
  year={2023},
  volume={abs/2302.11713},
  url={https://api.semanticscholar.org/CorpusID:257102433}
}

@article{Mensink2023EncyclopedicVV,
  title={Encyclopedic VQA: Visual questions about detailed properties of fine-grained categories},
  author={Thomas Mensink and Jasper R. R. Uijlings and Llu{\'i}s Castrej{\'o}n and Arushi Goel and Felipe Cadar and Howard Zhou and Fei Sha and Andre F. de Ara{\'u}jo and Vittorio Ferrari},
  journal={2023 IEEE/CVF International Conference on Computer Vision (ICCV)},
  year={2023},
  pages={3090-3101},
  url={https://api.semanticscholar.org/CorpusID:259165602}
}

@inproceedings{Deng2025EnablingCP,
  title={Enabling Collaborative Parametric Knowledge Calibration for Retrieval-Augmented Vision Question Answering},
  author={Jiaqi Deng and Kaize Shi and Zonghan Wu and Huan Huo and Dingxian Wang and Guandong Xu},
  year={2025},
  url={https://api.semanticscholar.org/CorpusID:277621934}
}

@article{Su2024SKVQASK,
  title={SK-VQA: Synthetic Knowledge Generation at Scale for Training Context-Augmented Multimodal LLMs},
  author={Xin Su and Man Luo and Kris W Pan and Tien Pei Chou and Vasudev Lal and Phillip Howard},
  journal={ArXiv},
  year={2024},
  volume={abs/2406.19593},
  url={https://api.semanticscholar.org/CorpusID:270845672}
}

@inproceedings{Bulian2022TomaytoTB,
  title={Tomayto, Tomahto. Beyond Token-level Answer Equivalence for Question Answering Evaluation},
  author={Jannis Bulian and Christian Buck and Wojciech Gajewski and Benjamin Boerschinger and Tal Schuster},
  booktitle={Conference on Empirical Methods in Natural Language Processing},
  year={2022},
  url={https://api.semanticscholar.org/CorpusID:246864012}
}

@article{Liu2023ImprovedBW,
  title={Improved Baselines with Visual Instruction Tuning},
  author={Haotian Liu and Chunyuan Li and Yuheng Li and Yong Jae Lee},
  journal={2024 IEEE/CVF Conference on Computer Vision and Pattern Recognition (CVPR)},
  year={2023},
  pages={26286-26296},
  url={https://api.semanticscholar.org/CorpusID:263672058}
}

@article{Qi2024RoRAVLMRR,
  title={RoRA-VLM: Robust Retrieval-Augmented Vision Language Models},
  author={Jingyuan Qi and Zhiyang Xu and Rulin Shao and Yang Chen and dingnan jin and Yu Cheng and Qifan Wang and Lifu Huang},
  journal={ArXiv},
  year={2024},
  volume={abs/2410.08876},
  url={https://api.semanticscholar.org/CorpusID:273323563}
}

@article{Caffagni2024WikiLLaVAHR,
  title={Wiki-LLaVA: Hierarchical Retrieval-Augmented Generation for Multimodal LLMs},
  author={Davide Caffagni and Federico Cocchi and Nicholas Moratelli and Sara Sarto and Marcella Cornia and Lorenzo Baraldi and Rita Cucchiara},
  journal={2024 IEEE/CVF Conference on Computer Vision and Pattern Recognition Workshops (CVPRW)},
  year={2024},
  pages={1818-1826},
  url={https://api.semanticscholar.org/CorpusID:269330022}
}

@inproceedings{Lerner2024CrossmodalRF,
  title={Cross-modal Retrieval for Knowledge-based Visual Question Answering},
  author={Paul Lerner and Olivier Ferret and Camille Guinaudeau},
  booktitle={European Conference on Information Retrieval},
  year={2024},
  url={https://api.semanticscholar.org/CorpusID:266933532}
}

@article{Bai2025Qwen25VLTR,
  title={Qwen2.5-VL Technical Report},
  author={Shuai Bai and Keqin Chen and Xuejing Liu and Jialin Wang and Wenbin Ge and Sibo Song and Kai Dang and Peng Wang and Shijie Wang and Jun Tang and Humen Zhong and Yuanzhi Zhu and Mingkun Yang and Zhaohai Li and Jianqiang Wan and Pengfei Wang and Wei Ding and Zheren Fu and Yiheng Xu and Jiabo Ye and Xi Zhang and Tianbao Xie and Zesen Cheng and Hang Zhang and Zhibo Yang and Haiyang Xu and Junyang Lin},
  journal={ArXiv},
  year={2025},
  volume={abs/2502.13923},
  url={https://api.semanticscholar.org/CorpusID:276449796}
}

@article{Sun2023EVACLIPIT,
  title={EVA-CLIP: Improved Training Techniques for CLIP at Scale},
  author={Quan Sun and Yuxin Fang and Ledell Yu Wu and Xinlong Wang and Yue Cao},
  journal={ArXiv},
  year={2023},
  volume={abs/2303.15389},
  url={https://api.semanticscholar.org/CorpusID:257766387}
}

@inproceedings{Chen2024M3EmbeddingMM,
  title={M3-Embedding: Multi-Linguality, Multi-Functionality, Multi-Granularity Text Embeddings Through Self-Knowledge Distillation},
  author={Jianlv Chen and Shitao Xiao and Peitian Zhang and Kun Luo and Defu Lian and Zheng Liu},
  booktitle={Annual Meeting of the Association for Computational Linguistics},
  year={2024},
  url={https://api.semanticscholar.org/CorpusID:267413218}
}

@article{DBLP:journals/corr/abs-2604-03647,
  author       = {Yunyao Yu and
                  Zhengxian Wu and
                  Zhuohong Chen and
                  Hangrui Xu and
                  Zirui Liao and
                  Xiangwen Deng and
                  Zhifang Liu and
                  Senyuan Shi and
                  Haoqian Wang},
  title        = {Stabilizing Unsupervised Self-Evolution of MLLMs via Continuous Softened
                  Retracing reSampling},
  journal      = {CoRR},
  volume       = {abs/2604.03647},
  year         = {2026},
  url          = {https://doi.org/10.48550/arXiv.2604.03647},
  doi          = {10.48550/ARXIV.2604.03647},
  eprinttype   = {arXiv},
  eprint       = {2604.03647},
  bibsource    = {dblp computer science bibliography, https://dblp.org}
}

@misc{xu2026visualsimilarityentityalignedretrieval,
      title={Beyond Visual Similarity: Entity-Aligned Retrieval for Knowledge-Based Visual Question Answering}, 
      author={Hangrui Xu and Zhengxian Wu and Yunyao Yu and Zhuohong Chen and Rui Cong and Xiangwen Deng and Zhifang Liu and Peng Jiao and Haoqian Wang},
      year={2026},
      eprint={2608.21450},
      archivePrefix={arXiv},
      primaryClass={cs.CV},
      url={https://arxiv.org/abs/2608.21450}, 
}

@article{wu2026promsa,
  title={ProMSA: Progressive Multimodal Search Agents for Knowledge-Based Visual Question Answering},
  author={Wu, ZhengXian and Xu, Hangrui and Shi, Kai and Chen, Zhuohong and Yu, Yunyao and Zhang, Chuanrui and Liao, Zirui and Yang, Jun and Yang, Zhenyu and Lu, Haonan and others},
  journal={arXiv preprint arXiv:2606.27974},
  year={2026}
}

@article{wu2026models,
  title={When Models Judge Themselves: Unsupervised Self-Evolution for Multimodal Reasoning},
  author={Wu, Zhengxian and Shi, Kai and Zhang, Chuanrui and Liao, Zirui and Yang, Jun and Yang, Ni and Peng, Qiuying and Zhang, Luyuan and Xu, Hangrui and Su, Tianhuang and others},
  journal={arXiv preprint arXiv:2603.21289},
  year={2026}
}

@inproceedings{DBLP:conf/aaai/WuZJXLZLJW26,
  author       = {Zhengxian Wu and
                  Chuanrui Zhang and
                  Shenao Jiang and
                  Hangrui Xu and
                  Zirui Liao and
                  Luyuan Zhang and
                  Huaqiu Li and
                  Peng Jiao and
                  Haoqian Wang},
  editor       = {Sven Koenig and
                  Chad Jenkins and
                  Matthew E. Taylor},
  title        = {Language-Guided and Motion-Aware Gait Representation for Generalizable
                  Recognition},
  booktitle    = {Fortieth {AAAI} Conference on Artificial Intelligence, Thirty-Eighth
                  Conference on Innovative Applications of Artificial Intelligence,
                  Sixteenth Symposium on Educational Advances in Artificial Intelligence,
                  {AAAI} 2026, Singapore, January 20-27, 2026},
  pages        = {10871--10878},
  publisher    = {{AAAI} Press},
  year         = {2026},
  url          = {https://doi.org/10.1609/aaai.v40i13.38063},
  doi          = {10.1609/AAAI.V40I13.38063},
  bibsource    = {dblp computer science bibliography, https://dblp.org}
}

@article{DBLP:journals/corr/abs-2602-00104,
  author       = {Zhuohong Chen and
                  Zhengxian Wu and
                  Zirui Liao and
                  Shenao Jiang and
                  Hangrui Xu and
                  Yang Chen and
                  Chaokui Su and
                  Xiaoyu Liu and
                  Haoqian Wang},
  title        = {{R3G:} {A} Reasoning-Retrieval-Reranking Framework for Vision-Centric
                  Answer Generation},
  journal      = {CoRR},
  volume       = {abs/2602.00104},
  year         = {2026},
  url          = {https://doi.org/10.48550/arXiv.2602.00104},
  doi          = {10.48550/ARXIV.2602.00104},
  eprinttype   = {arXiv},
  eprint       = {2602.00104},
  bibsource    = {dblp computer science bibliography, https://dblp.org}
}

@inproceedings{Liu2026OVIBenchBO,
  title={OVIBench: Benchmarking Online Video Question Answering under Interruption},
  author={Naiming Liu and Zhiheng Wu and Shuning Wang and Tie Zhang and Bowen Liu and Tong Wang},
  year={2026},
  url={https://api.semanticscholar.org/CorpusID:291328938}
}

\appendix

\section{Additional Experimental Analyses}
\label{sec:additional_experiments}

This section provides additional analyses to better understand the behavior of DBAgent. We include analyses on retrieval-answer correlation, backbone generalization, inference cost, trajectory quality, tool-use behavior, direct-answer control, and difficulty sampling.

\subsection{Generalization across Backbones}
\label{sec:backbone_generalization}

The main experiments use Qwen2.5-VL-7B as the backbone. To examine whether the proposed decision-based training generalizes beyond this model, we further evaluate DBAgent with Qwen2.5-VL-3B-Instruct and InternVL3-8B-Instruct. Qwen2.5-VL-3B tests a smaller model from the same family, while InternVL3-8B tests a different VLM family with a comparable scale.

\begin{table*}[t]
\centering
\small
\setlength{\tabcolsep}{4.5pt}
\renewcommand{\arraystretch}{1.12}
\resizebox{\textwidth}{!}{%
\begin{tabular}{l l c c c c c}
\toprule
Method & Backbone & E-VQA Single-Hop & E-VQA All & InfoSeek Unseen-Q & InfoSeek Unseen-E & InfoSeek All \\
\midrule
DBAgent w/o SFT & Qwen2.5-VL-3B & 20.4 & 19.6 & 20.5 & 21.8 & 21.2 \\
DBAgent SFT Mixed & Qwen2.5-VL-3B & 41.3 & 39.8 & 39.5 & 44.7 & 42.2 \\
DBAgent w/o SFT & InternVL3-8B & 25.8 & 25.1 & 25.0 & 25.7 & 25.4 \\
DBAgent SFT Mixed & InternVL3-8B & 45.6 & 44.3 & 43.1 & 49.2 & 46.5 \\
DBAgent SFT Mixed & Qwen2.5-VL-7B & \textbf{46.0} & \textbf{45.2} & \textbf{43.6} & \textbf{50.4} & \textbf{48.7} \\
\bottomrule
\end{tabular}
}
\caption{Generalization of DBAgent across different backbone models.}
\label{tab:backbone_generalization}
\end{table*}

As shown in Table~\ref{tab:backbone_generalization}, trajectory SFT consistently improves over the corresponding w/o-SFT variant for both Qwen2.5-VL-3B and InternVL3-8B. This suggests that DBAgent is not tied to a single backbone. Instead, the trajectory-based decision training transfers to a smaller model in the same family and to a different VLM family, supporting the generality of the proposed framework.

\subsection{Inference Cost}
\label{sec:inference_cost}

Because DBAgent performs iterative decision making and tool use, it introduces additional inference cost compared with single-pass models. We therefore report a relative efficiency comparison in Table~\ref{tab:inference_cost}. The relative time is normalized by the inference time of Qwen2.5-VL-7B Base.

\begin{table}[t]
\centering
\small
\setlength{\tabcolsep}{6pt}
\renewcommand{\arraystretch}{1.12}
\resizebox{\columnwidth}{!}{%
\begin{tabular}{l c c}
\toprule
Method & Relative Time & Avg. Accuracy \\
\midrule
Qwen2.5-VL-7B Base & 1.0$\times$ & 22.0 \\
Wiki-LLaVA & 1.4$\times$ & 24.6 \\
VLM-PRF (Qwen2.5-VL-7B) & 1.6$\times$ & 39.4 \\
ReflectiVA (Qwen2.5-VL-7B) & 1.9$\times$ & 40.4 \\
DBAgent SFT Mixed & 2.1$\times$ & \textbf{47.0} \\
\bottomrule
\end{tabular}
}
\caption{Relative inference cost and average accuracy across E-VQA and InfoSeek.}
\label{tab:inference_cost}
\end{table}

DBAgent has higher inference cost due to multi-step decisions and retrieval calls. However, the overhead remains moderate under the predefined tool budget, and the model obtains the best average accuracy among the compared methods. This result characterizes the trade-off between adaptive search and inference efficiency.

\subsection{Trajectory Quality}
\label{sec:trajectory_quality}

To examine whether trajectory SFT improves evidence-grounded behavior rather than only output-format compliance, we evaluate the supportiveness of generated trajectories using GPT-4o as an automatic judge. For correctly answered samples, the judge is given the question, generated trajectory, retrieved evidence, and final answer, and then determines whether the trajectory evidence supports the answer.

\begin{table}[t]
\centering
\small
\setlength{\tabcolsep}{5pt}
\renewcommand{\arraystretch}{1.12}
\resizebox{\columnwidth}{!}{%
\begin{tabular}{l l c c}
\toprule
Dataset & Model & Answer Acc. & Support Rate \\
\midrule
E-VQA & DBAgent w/o SFT & 23.9 & 42.7 \\
E-VQA & DBAgent SFT Mixed & \textbf{45.2} & \textbf{84.8} \\
InfoSeek & DBAgent w/o SFT & 24.4 & 43.4 \\
InfoSeek & DBAgent SFT Mixed & \textbf{48.7} & \textbf{85.6} \\
\bottomrule
\end{tabular}
}
\caption{GPT-4o judge evaluation of answer-supporting trajectories.}
\label{tab:trajectory_quality}
\end{table}

Table~\ref{tab:trajectory_quality} shows that SFT substantially increases the answer support rate on both datasets. This indicates that trajectory supervision helps the model generate evidence-grounded decision paths, not merely well-formatted action tags. The improvement in support rate also aligns with the gain in answer accuracy, suggesting that better trajectories contribute to more reliable final predictions.

\subsection{Relationship Between Retrieval and Answer Correctness}
\label{sec:retrieval_answer_analysis}

To better understand the behavior of DBAgent, we analyze the relationship between retrieval correctness and final answer correctness. As shown in Table~\ref{tab:retrieval_answer_2x2}, correct retrieval is strongly associated with correct answering: 70.4\% of retrieval-correct samples are answered correctly, whereas 88.6\% of retrieval-incorrect samples lead to wrong answers. Meanwhile, the two off-diagonal cases show that retrieval and answering are not perfectly coupled. Some samples can still be answered when retrieval is incorrect, while some retrieval-correct samples fail due to reasoning or evidence-use errors. These results indicate that correct retrieval is important but not sufficient, further motivating a decision-based agent that can adaptively retrieve, refine queries, and decide when to stop.

\begin{table}[t]
\centering
\small
\begin{adjustbox}{width=\columnwidth}
\begin{tabular}{lcc}
\toprule
 & \textbf{Answer Correct} & \textbf{Answer Wrong} \\
\midrule
\textbf{Retrieval Correct}   & 70.4 & 29.6 \\
\textbf{Retrieval Incorrect} & 11.4 & 88.6 \\
\bottomrule
\end{tabular}
\end{adjustbox}
\caption{Relationship between retrieval correctness and answer correctness. We report the proportion (\%) of samples in each case.}
\label{tab:retrieval_answer_2x2}
\end{table}

\subsection{Tool-Use Behavior}
\label{sec:tool_use_behavior}

We further compare the tool-use behavior before and after trajectory SFT. This analysis helps determine whether the model actually changes its decision policy after training.

\begin{table*}[t]
\centering
\small
\setlength{\tabcolsep}{5pt}
\renewcommand{\arraystretch}{1.12}
\resizebox{\textwidth}{!}{%
\begin{tabular}{l c c c c c}
\toprule
Model & Direct Answer Ratio & Image Search Used & Text Search Used & Multi-step Retrieval & Avg. Retrieval Calls \\
\midrule
DBAgent w/o SFT & 64.8 & 23.5 & 14.9 & 4.9 & 0.40 \\
DBAgent SFT Mixed & 5.4 & \textbf{41.4} & \textbf{68.9} & \textbf{32.8} & \textbf{1.27} \\
\bottomrule
\end{tabular}
}
\caption{Tool-use behavior before and after trajectory SFT. Ratios are reported in percent except for average retrieval calls. The usage ratios are not mutually exclusive because a trajectory may use both image and text search.}
\label{tab:tool_use_behavior}
\end{table*}

As shown in Table~\ref{tab:tool_use_behavior}, the w/o-SFT model frequently falls back to direct answering, while the SFT model uses image search, text search, and multi-step retrieval more often. This supports our claim that trajectory SFT changes the model's decision behavior and teaches a state-conditioned decision policy instead of simply adding an agent-style interface.

\subsection{Direct-Answer Control}
\label{sec:direct_answer_control}

The retrieval-answer analysis in Table~\ref{tab:retrieval_answer_2x2} shows a strong correlation between retrieval correctness and answer correctness. To check whether this relationship is mainly driven by questions answerable from parametric knowledge, we remove retrieved evidence from samples where both retrieval and final answer are correct, and then ask the same Qwen2.5-VL-7B backbone to answer using only the image and question.

\begin{table}[t]
\centering
\small
\setlength{\tabcolsep}{8pt}
\renewcommand{\arraystretch}{1.12}
\begin{tabular}{l c}
\toprule
Dataset & Direct Answer w/o Retrieval \\
\midrule
E-VQA & 32.8 \\
InfoSeek & 35.1 \\
\bottomrule
\end{tabular}
\caption{Direct-answer control on samples with correct retrieval and correct final answers.}
\label{tab:direct_answer_control}
\end{table}

Table~\ref{tab:direct_answer_control} shows that accuracy drops substantially after removing retrieval evidence. Therefore, the correct-retrieval and correct-answer cases are not mainly solved by parametric knowledge; they rely on useful external evidence. This further supports the interpretation that retrieval correctness is a meaningful factor in final answer quality.

\subsection{Difficulty Sampling Ratio}
\label{sec:sampling_ratio}

Finally, we evaluate the effect of different easy:medium:hard sampling ratios used for trajectory construction. The goal of balanced sampling is to expose the model to diverse decision states and avoid bias toward direct answering, single-step retrieval, or over-retrieval.

\begin{table}[t]
\centering
\small
\setlength{\tabcolsep}{8pt}
\renewcommand{\arraystretch}{1.12}
\begin{tabular}{l c c}
\toprule
Sampling Ratio & E-VQA All & InfoSeek All \\
\midrule
2:1:1 & 43.7 & 46.2 \\
1:2:1 & 44.1 & 46.8 \\
1:1:2 & 44.5 & 47.1 \\
1:1:1 & \textbf{45.2} & \textbf{48.7} \\
\bottomrule
\end{tabular}
\caption{Ablation on the easy:medium:hard difficulty sampling ratio.}
\label{tab:sampling_ratio}
\end{table}

The results in Table~\ref{tab:sampling_ratio} show that the balanced 1:1:1 ratio performs best on both benchmarks. This suggests that balanced trajectory construction helps avoid both direct-answer collapse and unnecessary over-retrieval, leading to more robust decision learning.

\section{Datasets}

We evaluate our method on two widely used knowledge-based visual question answering benchmarks, InfoSeek and E-VQA. Both datasets are designed to test whether a model can go beyond the visible content of an image and incorporate external knowledge. However, they differ significantly in their construction principles, knowledge types, and reasoning requirements. This diversity allows us to comprehensively examine the decision-making and search behaviors of our agent.

\paragraph{InfoSeek.}
InfoSeek is a large-scale benchmark specifically designed for knowledge-intensive visual question answering. Each example consists of an image, a natural language question, and a set of acceptable answers. Unlike traditional VQA datasets that focus on object recognition or simple visual attributes, InfoSeek emphasizes long-tail entities and factual knowledge that is rarely memorized by vision-language models.

A key characteristic of InfoSeek is that most questions cannot be answered solely based on the image. Instead, the image serves as a visual anchor that points to a specific entity, such as a landmark, an animal species, a historical artifact, or a cultural object. The question then asks for factual attributes of that entity, such as its origin, function, scientific classification, or historical background. Answering such questions typically requires consulting an external knowledge source.

InfoSeek also exhibits strong diversity in question types. Some questions can be resolved with a single retrieval step, while others require multiple rounds of refinement because the initial query is too ambiguous or incomplete. This makes InfoSeek particularly suitable for evaluating whether a model can decide when to retrieve, how to formulate queries, and how to revise them based on newly obtained evidence.

\paragraph{E-VQA.}
E-VQA is another benchmark designed for entity-centric visual question answering. Each question is explicitly tied to a visual entity in the image, and the answer usually involves factual or encyclopedic knowledge about that entity. Compared to InfoSeek, E-VQA focuses more on fine-grained attributes, such as taxonomic categories, functional properties, or specific biographical facts.

A notable feature of E-VQA is that many questions assume the model can correctly identify the entity from the image. However, recognizing the entity alone is often insufficient to answer the question. The model must also retrieve or recall the corresponding factual information. This makes E-VQA a good testbed for separating two distinct challenges: visual grounding and knowledge acquisition.

In addition, E-VQA includes a wide range of domains, such as animals, plants, monuments, tools, and artworks. This diversity further increases the difficulty of relying on parametric knowledge alone, especially for long-tail concepts.

\paragraph{Comparison and Motivation.}
Although both datasets require external knowledge, they differ in how that knowledge is triggered. In InfoSeek, the model often needs to discover what the image refers to before it can even formulate a meaningful query. In E-VQA, the model can often identify the entity but still lacks the specific attribute required by the question.

These differences make the two datasets complementary. InfoSeek stresses the need for adaptive entity discovery and query formulation, while E-VQA emphasizes fine-grained factual lookup after visual grounding. Together, they form a challenging testbed for dynamic, multi-step reasoning.

Importantly, neither dataset can be reliably solved with a fixed retrieval-then-answering pipeline. Some questions require no retrieval at all, some require a single retrieval step, and others require multiple refinements. This variability motivates our formulation of KB-VQA as a decision-making process rather than a static pipeline. Our search-agent framework is designed to explicitly model this variability, enabling the agent to decide when to retrieve, which modality to use, and how to refine its queries.

\section{Baselines}

We compare our method with a broad set of baselines that represent different paradigms for multimodal reasoning and retrieval in KB-VQA. These baselines can be roughly grouped into three categories: zero-shot multimodal large language models, classical retrieval-augmented models with fixed pipelines, and retrieval-augmented models with explicit reasoning or reflection mechanisms. Together, they cover the most common design choices adopted in existing systems.

\subsection{Zero-shot Multimodal Large Language Models.}
We include several strong pretrained multimodal models that directly generate answers from images and questions without explicit access to external knowledge sources, including BLIP-2, InstructBLIP, LLaVA, and GPT-4V.

\paragraph{BLIP-2 and InstructBLIP.} These models connect a frozen large language model with a visual encoder via a lightweight query transformer. This design enables flexible multimodal generation while preserving the generalization ability of large language models. However, factual knowledge is implicitly stored in the model parameters, which makes these models sensitive to the coverage of the pretraining data. As a result, they often struggle with long-tail entities and rare facts.

\paragraph{LLaVA-v1.5.\cite{Liu2023ImprovedBW}} LLaVA follows a similar philosophy but is trained with large-scale multimodal instruction tuning. This improves its robustness and instruction-following ability, especially for conversational and compositional queries. Nevertheless, it still relies on parametric memory for factual information and does not explicitly interact with external knowledge bases.

\paragraph{GPT-4V.} GPT-4V is a strong proprietary vision-language model with broad general knowledge and strong reasoning ability. It serves as an important reference point for zero-shot performance. Despite its strength, its access to knowledge remains implicit, and it does not expose explicit retrieval or evidence-grounding behaviors.

\paragraph{Qwen2.5-VL.\cite{Bai2025Qwen25VLTR}} Qwen2.5-VL is a recent open-source multimodal large language model that supports strong visual understanding and instruction following. It integrates a vision encoder with a large language model and is trained with large-scale multimodal instruction data. This design enables the model to handle a wide range of multimodal tasks, including visual question answering and image-grounded reasoning. However, similar to other zero-shot multimodal models, factual knowledge is mainly stored in model parameters. Without explicit retrieval mechanisms, the model may still struggle with questions that require external knowledge or long-tail factual information.

\subsection{Classical Retrieval-Augmented Models.}

\paragraph{DPR.\cite{Lerner2024CrossmodalRF}} DPR is a dense passage retrieval framework originally developed for open-domain question answering. It uses a dual-encoder architecture to embed queries and passages into a shared vector space. In our setting, DPR serves as a representative text retriever that retrieves relevant passages based on a fixed query formulation.

\paragraph{RORA-VLM.\cite{Qi2024RoRAVLMRR}} RORA-VLM integrates retrieval into the multimodal pipeline by combining a vision-language model with a CLIP-based image retriever. Retrieved visual content is used to augment the generation process, allowing the model to access additional visual context beyond the input image.

\paragraph{EchoSight.\cite{Yan2024EchoSightAV}} EchoSight is a multimodal retrieval-augmented framework that retrieves visually or semantically related images and associated textual descriptions using CLIP-style encoders. The retrieved information is concatenated with the original input and passed to a language model for answer generation.

\paragraph{Wiki-LLaVA.\cite{Caffagni2024WikiLLaVAHR}} Wiki-LLaVA augments a vision-language model with external Wikipedia knowledge. It retrieves relevant articles or passages using a CLIP-based retriever and fuses the retrieved content into the language model input.

\subsection{Retrieval-Augmented Models with Reasoning.}

\paragraph{ReflectiVA.\cite{Cocchi2024AugmentingML}} ReflectiVA introduces a reflection-based reasoning strategy that encourages the model to generate intermediate reasoning traces and revise its answer if inconsistencies are detected. This design improves robustness and reduces certain types of hallucinations.

\paragraph{VL-MPRF.\cite{Hong2025KnowledgebasedVQ}} VL-MPRF incorporates a multi-perspective reasoning framework that aggregates retrieved evidence from different views. It emphasizes structured reasoning over multiple knowledge snippets and integrates the reasoning process into the language model through prompt-based decomposition.

\paragraph{Unified Discussion.}
Although the above methods differ in architecture and training strategies, they share a common assumption: the structure of the reasoning and retrieval process is largely predefined. Retrieval is either always performed or implicitly triggered by prompt templates. The model is not explicitly trained to decide whether retrieval is necessary, which modality to use, or when to stop.

Moreover, query formulation is often treated as a static transformation rather than a learnable decision. When retrieval fails or returns insufficient information, most systems either hallucinate or passively consume the returned evidence.
In practice, KB-VQA exhibits diverse difficulty levels. Some questions can be answered directly from the image, while others require one or multiple rounds of refinement. A fixed pipeline cannot naturally express this diversity.

\section{Retriever}
\paragraph{EVA-CLIP.\cite{Sun2023EVACLIPIT}} 
EVA-CLIP is a large-scale vision-language representation model built upon the CLIP framework. It improves visual representation quality by scaling both the vision encoder and the pretraining data. The model is trained with extensive image-text pairs and adopts stronger visual backbones, which leads to improved cross-modal alignment between images and text. In our framework, EVA-CLIP is used as the image retriever to retrieve visually or semantically related images from the external knowledge base. The retrieved images provide additional visual context that can support subsequent reasoning and answer generation.

\paragraph{BGE-M3.\cite{Chen2024M3EmbeddingMM}} 
BGE-M3 is a multilingual dense retrieval model designed for general-purpose information retrieval. It produces high-quality text embeddings that support both semantic matching and cross-lingual retrieval. Compared with earlier dense retrievers, BGE-M3 demonstrates strong performance across various retrieval benchmarks and supports flexible query formulations. In our framework, BGE-M3 is used as the text retriever to retrieve relevant textual evidence from the knowledge base. The retrieved passages provide complementary factual information that may not be directly inferred from the visual content alone.

\section{DBAgent Inference Algorithm}
\label{sec:appendix}
This appendix provides the detailed inference procedure of DBAgent, corresponding to the decision process described in Section \ref{sec:rollout}. During inference, the agent iteratively generates reasoning tokens and selects actions based on the current information state. Depending on the generated action tag, the system invokes different retrieval tools and appends the returned evidence to the state. The process continues until the agent outputs the final answer or reaches the maximum turn budget.

\begin{algorithm}[H]
\caption{Inference procedure of DBAgent with multi-turn tool interaction.}
\label{alg:rollout}
\textbf{Require:} image $x$, question $q$, policy model $\pi_{\theta}$, text retriever $\mathcal{R}_{\text{text}}$, image retriever $\mathcal{R}_{\text{img}}$, maximum turn budget $B$ \\
\textbf{Ensure:} final answer $y$
\begin{algorithmic}[1]
\raggedright
\State Initialize output $y \gets \emptyset$, evidence buffer $\mathcal{E} \gets \emptyset$
\State Initialize turn counter $b \gets 0$
\While{$b < B$}
    \State Initialize one-turn output $y_b \gets \emptyset$
    \While{true}
        \State Sample next token $z \sim \pi_{\theta}(\cdot \mid x, q, \mathcal{E}, y, y_b)$
        \State Append $z$ to $y_b$
        \If{$z$ is a closing tag or $\langle eos \rangle$}
            \State \textbf{break}
        \EndIf
    \EndWhile
    \State $y \gets y \Vert y_b$
    
    \If{\tagSearch{<text\_search>} is detected in $y_b$}
        \State Parse query $u$ from $y_b$
        \State Retrieve text evidence $e \gets \mathcal{R}_{\text{text}}(u)$
        \State Insert $\tagEv{<evidence>}~e~\tagEv{</evidence>}$ into $\mathcal{E}$
        
    \ElsIf{\tagSearch{<image\_search>} is detected in $y_b$}
        \State Parse image input $q_{\text{img}}$ from $y_b$
        \If{$q_{\text{img}}$ is missing}
            \State $q_{\text{img}} \gets x$
        \EndIf
        \State Retrieve visual evidence $e \gets \mathcal{R}_{\text{img}}(q_{\text{img}})$
        \State Insert $\tagEv{<evidence>}~e~\tagEv{</evidence>}$ into $\mathcal{E}$
        
    \ElsIf{\tagAnswer{<answer>} is detected in $y_b$}
        \State Parse final answer span $y^\star$
        \State \Return $y^\star$
        
    \Else
        \State Optionally append a short reflection message
    \EndIf
    
    \State $b \gets b + 1$
\EndWhile
\State \Return the last valid answer span if any
\end{algorithmic}
\end{algorithm}
% --------------------------------------------------------

\section{Training Details}

\paragraph{Training setup.}
We fine-tune our search-agent model using full-parameter supervised fine-tuning on the constructed multi-stage trajectories. All experiments are conducted on 8 NVIDIA A800 GPUs, each with 80GB memory. To support long multi-turn contexts and reduce memory consumption, we enable bfloat16 precision, gradient checkpointing, and ZeRO-3 optimization.

\paragraph{Optimization and scheduling.}
We adopt Qwen2.5-VL-7B-Instruct as the backbone model. The per-device training batch size is set to 1, and we use gradient accumulation with 4 steps, resulting in a global batch size of 32. We use a cosine learning rate scheduler with an initial learning rate of $7 \times 10^{-6}$ and a warmup ratio of 0.06. Weight decay is set to 0.

\paragraph{Sequence length and training duration.}
To accommodate long reasoning trajectories, we set the maximum sequence length to 16,384 tokens. The model is trained for 3 epochs over approximately 200k trajectories. We randomly hold out 2\% of the training data as a validation set and perform evaluation every 2000 steps. We report the best-performing checkpoint based on validation accuracy. All models are trained using the same optimization settings across datasets to ensure fair comparison.

\paragraph{Reproducibility.}
All hyperparameters, prompts, and trajectory construction rules are provided in the supplementary material. Our training setup follows standard practices in large-scale multimodal instruction tuning and can be reproduced with commonly available GPU clusters.

\section{Prompt Design}

This subsection describes the prompts used in our search-agent framework. They cover two settings: (1) the main agent prompt used for inference and training, and (2) a set of multi-stage prompts used to construct high-quality trajectories. Our goal is not only to teach the model to answer questions, but also to teach it when external knowledge is needed, which tool to use, how to write better queries, and when to stop.

Unlike retrieval-augmented generation pipelines that follow a fixed order of steps, we treat KB-VQA as a multi-step decision process. At each step, the model reasons about the current state, selects an action, and updates its decision after observing new information. The prompts are designed to expose these intermediate decisions in a standardized, machine-readable format.

\paragraph{Search-agent prompt for inference and training.}
The core prompt follows a simple think--act pattern. At each turn, the model first writes a short reasoning trace inside \texttt{<think>}, then outputs exactly one action tag. This structure encourages the model to reflect when it receives new information, instead of retrieving by default.

\paragraph{Answer action.}
The model outputs \texttt{<answer>} when the current image and available evidence are sufficient. The answer must be short and contain only the final answer span.

\paragraph{Text-retrieval action.}
The model outputs \texttt{<text\_search>} when it cannot answer yet but can identify what the question refers to in the image. The query must be concise and specific. It should combine an entity name (or a short visual description) with the attribute asked by the question, so the retriever can fetch the missing knowledge efficiently.

\paragraph{Image-retrieval action.}
The model outputs \texttt{<image\_search>} when it cannot answer and also cannot reliably identify the key entity from the image. To keep a stable tool interface, the content of this tag is fixed to the placeholder token \texttt{image\_path}.

\paragraph{Optional caption for query refinement.}
If the model has already performed image retrieval and still needs text retrieval, it may output an optional \texttt{<caption>} block before the final action. The caption must describe only what is visible, without speculation. Its role is to provide concrete visual keywords that help form a better text query.

\paragraph{Formatting constraints.}
We enforce strict formatting rules. Each turn must start with \texttt{<think>} and end with exactly one action tag. Only the allowed tags may appear, and no trailing text is permitted. This makes the reasoning, the decision, and the tool usage explicit and easy to supervise.

\paragraph{Multi-stage trajectory construction.}
To teach dynamic decision-making, we construct trajectories with multiple stages. Each stage uses two prompts: an answering prompt and a judging prompt. The answering prompt produces an answer given the currently available information. The judging prompt then checks correctness (offline) and, when the answer is wrong, decides what to do next and produces a rewritten query or a tool choice. This design allows us to build trajectories with different structures, including direct answering, single-step retrieval, and multi-step refinement.

\paragraph{Stage 1: initial reasoning and routing.}
Stage 1 uses only the image and the question. The answering prompt requests a reasoning trace, the main entity, and a tentative answer. The judging prompt verifies the answer. If it is wrong, the judge determines whether the failure is caused by incorrect entity grounding or missing knowledge. If the entity is wrong, it routes to image retrieval. If the entity is correct but knowledge is missing, it routes to text retrieval. In both cases, it outputs the next action in a standardized format.

\paragraph{Stage 2: tool-based re-answering and query rewriting.}
Stage 2 follows the routing decision from Stage 1. In the image-based branch, the model receives text evidence obtained through image retrieval and answers again using the image plus evidence. If the answer is still unsupported, the judging prompt produces a detailed visual caption and a rewritten text query, where the caption is restricted to visible content only. In the text-based branch, the model answers using retrieved text evidence. If the answer remains unsupported, the judge explains what information is missing and outputs a new text query that is meaningfully different from the previous one.

\paragraph{Stage 3: final re-answering with new evidence.}
Stage 3 handles harder cases where Stage 2 is still insufficient. The model is given the full history of the trajectory together with newly retrieved text evidence from an updated query. The goal is to treat the new evidence as the primary factual source and produce a final answer. The Stage-3 answering prompt follows the same contract as Stage 2: a short \texttt{<think>} block focusing on what matters in the image and how the new evidence supports the conclusion, followed by a final \texttt{<answer>} block. No other tags are allowed.

\paragraph{Offline judging for data construction.}
For trajectory construction, we also use judging prompts to label whether the answer at each stage is correct under a semantic match rule (case-insensitive, minor wording differences allowed). When an answer is wrong, the judge outputs the next action choice and a rewritten query in the required format. To keep the trace realistic, the judge is instructed not to mention gold answers or evaluation language in its explanations.

\paragraph{Design principles.}
We follow a few simple principles. First, decisions are explicit: the model must state its reasoning and select actions directly. Second, the output format is tightly constrained, which makes trajectories easy to parse and suitable for supervised learning. Third, answering and judging are separated, so we can supervise both solving and decision-making. Fourth, failed attempts are treated as useful signal: instead of discarding them, we turn them into refinement steps that teach better tool use. Finally, the framework is fully multimodal, because the image is considered at every stage rather than only at the beginning.

\begin{tcolorbox}[
  promptboxstyle,
  colback=gray!10!white,
  colframe=gray!50!black,
  title={Search-Agent Prompt (Inference \& Training)}
]

\textbf{Task.} Answer the question. Always reason in
\texttt{<think>}...\texttt{</think>} when you get new information
(image or \texttt{<evidence>}...\texttt{</evidence>}).

\vspace{2mm}
\textbf{Choose EXACTLY ONE action per turn:}

\vspace{1mm}
\textbf{Action 1: Answer.}

{\ttfamily\footnotesize
<answer> ... </answer>
}

Use this action if the current image/evidence is sufficient to answer exactly.

\vspace{1.5mm}
\textbf{Action 2: Text Retrieval.}

{\ttfamily\footnotesize
<text\_search> QUERY </text\_search>
}

Use this action if you cannot answer yet but you can identify what the question refers to in the image.

\textbf{QUERY rule:}
\begin{itemize}[leftmargin=4mm, topsep=1pt, itemsep=0pt]
  \item QUERY = (entity name or short visual description) + (asked attribute)
  \item Keep it concise and specific.
\end{itemize}

\vspace{1.5mm}
\textbf{Action 3: Image Retrieval.}

{\ttfamily\footnotesize
<image\_search>image\_path</image\_search>
}

Use this action if you cannot answer and you cannot identify the key entity from the image.

\vspace{1.5mm}
\textbf{IMPORTANT:}
\begin{itemize}[leftmargin=4mm, topsep=1pt, itemsep=0pt]
  \item The content inside \texttt{<image\_search>} MUST be exactly \texttt{image\_path} (fixed placeholder).
  \item Do NOT change it.
\end{itemize}

\vspace{1.5mm}
\textbf{Caption (optional):}

If you already used \texttt{<image\_search>} and still need \texttt{<text\_search>}, output an optional caption to help write a better query.

{\ttfamily\footnotesize
<caption> ... </caption>
}

\begin{itemize}[leftmargin=4mm, topsep=1pt, itemsep=0pt]
  \item Caption must be ONE short visible-only sentence.
  \item \texttt{<caption>} may appear only before the final action tag.
\end{itemize}

\vspace{1.5mm}
\textbf{Format rules:}
\begin{itemize}[leftmargin=4mm, topsep=1pt, itemsep=0pt]
  \item Output must start with \texttt{<think>}.
  \item Only use tags: \texttt{<think>}, \texttt{<answer>}, \texttt{<text\_search>}, \texttt{<image\_search>}, \texttt{<caption>}.
  \item End with exactly one action tag: \texttt{<answer>} or \texttt{<text\_search>} or \texttt{<image\_search>}.
  \item \texttt{<caption>} may appear only before the final action tag.
  \item After \texttt{</answer>}, \texttt{</text\_search>}, or \texttt{</image\_search>}, output nothing.
\end{itemize}

\vspace{1.5mm}
\textbf{Answer format:}
\begin{itemize}[leftmargin=4mm, topsep=1pt, itemsep=0pt]
  \item \texttt{<answer>} must contain ONLY the final answer span.
  \item Keep it very short: prefer 1--4 words (or a single number/date/unit if required).
  \item Do NOT write a full sentence.
  \item Do NOT add explanations, punctuation, or prefixes.
\end{itemize}

\end{tcolorbox}

\begin{tcolorbox}[
  colback=blue!10!white,
  colframe=blue!50!black,
  title=\textbf{Stage-1 Multimodal VQA Prompt},
  fonttitle=\bfseries,
  fontupper=\small,
  sharp corners,
  breakable,
  left=2mm,right=2mm,top=1mm,bottom=1mm,
  before upper=\setlength{\parindent}{0pt}
]

You are a multimodal question answering model.
The user will provide one image and one question about that image.

\vspace{1.5mm}
\textbf{Your task consists of three steps:}

\vspace{1mm}
\textbf{Step 1: Reasoning.}

In \texttt{<think>...</think>}, write your detailed reasoning.

\begin{itemize}
  \setlength{\itemsep}{0pt}
  \setlength{\topsep}{2pt}
  \item Describe what you see in the image.
  \item Explain how the visual content relates to the question.
  \item Show the logical steps used to derive the answer.
  \item Do not include the final answer here.
\end{itemize}

\vspace{1mm}
\textbf{Step 2: Entity Identification.}

In \texttt{<entity>...</entity>}, output the main entity in the image.

\begin{itemize}
  \setlength{\itemsep}{0pt}
  \setlength{\topsep}{2pt}
  \item Use a short noun phrase.
  \item Output exactly one entity.
\end{itemize}

\vspace{1mm}
\textbf{Step 3: Final Answer.}

In \texttt{<answer>...</answer>}, output the final short answer only.

\vspace{1.5mm}
\textbf{Strict Output Format (Mandatory):}

{\ttfamily\small
<think>\dots</think>\par
<entity>\dots</entity>\par
<answer>\dots</answer>
}

\begin{itemize}
  \setlength{\itemsep}{0pt}
  \setlength{\topsep}{2pt}
  \item Each tag must appear exactly once and be properly closed.
  \item Do not output any other tags.
  \item Do not output anything before \texttt{<think>} or after \texttt{</answer>}.
  \item Do not write ``unknown'' in \texttt{<answer>}.
\end{itemize}

\vspace{1.5mm}
\textbf{Example Output Format}

{\ttfamily\small
<think>\par
I first look at the image. It shows a football player wearing a light blue and white striped jersey with the number 10. He is on a football pitch with a ball at his feet. This jersey pattern is characteristic of the Argentina national team, so the answer is Argentina national team.\par
</think>\par
<entity>Lionel Messi</entity>\par
<answer>Argentina national team</answer>
}

\end{tcolorbox}

% Preamble requirement:
% \usepackage{seqsplit}

\begin{tcolorbox}[
  colback=yellow!10!white,
  colframe=yellow!50!black,
  title=\textbf{Stage-1 Judgment Prompt},
  fonttitle=\bfseries,
  fontupper=\small,
  sharp corners,
  breakable,
  left=2mm,right=2mm,top=1mm,bottom=1mm,
  before upper=\setlength{\parindent}{0pt}
]

You are a strong language model used to \textbf{JUDGE and POST-PROCESS} Stage-1 multimodal QA outputs.

For each sample, you will receive the following fields:

\begin{itemize}
  \setlength{\itemsep}{0pt}
  \setlength{\topsep}{2pt}
  \item \texttt{[question]}: the question about the image
  \item \texttt{[stage1\_think]}: the reasoning text generated in Stage-1
  \item \texttt{[stage1\_entity]}: the entity predicted in Stage-1
  \item \texttt{[stage1\_answer]}: the answer predicted in Stage1
  \item \texttt{[gold\_answer]}: the ground-truth answer
  \item \texttt{[gold\_entity]}: the ground-truth entity (e.g., Wikipedia title)
\end{itemize}

\textbf{Interpreting \texttt{gold\_answer}:}
\vspace{-2mm}
\begin{itemize}
  \setlength{\itemsep}{0pt}
  \setlength{\topsep}{2pt}
  \item \texttt{gold\_answer} denotes a \textbf{set} of acceptable answers, where elements are separated by ``|''.
  \item \texttt{stage1\_answer} is considered \textbf{CORRECT} if it semantically matches \emph{any} element in this set.
\end{itemize}

\textbf{Your task consists of three steps:}

\vspace{1mm}
\textbf{Step 1: Answer Judgment.}

Decide whether \texttt{stage1\_answer} is \textbf{CORRECT} compared to \texttt{gold\_answer}.
\vspace{-2mm}
\begin{itemize}
  \setlength{\itemsep}{0pt}
  \setlength{\topsep}{2pt}
  \item Use semantic matching.
  \item Ignore case, punctuation, and minor wording differences.
\end{itemize}

\textbf{Step 2: Correct Case.}

If the answer is correct, output exactly:

{\ttfamily\small
[correct]
}and \textbf{nothing else}.

\vspace{2mm}
\textbf{Step 3: Wrong Case.}

If the answer is wrong, further judge whether \texttt{stage1\_entity} matches \texttt{gold\_entity}.

\vspace{2mm}
\textbf{Case (a): Entity is Wrong.}
\vspace{-2mm}
\begin{itemize}
  \setlength{\itemsep}{0pt}
  \setlength{\topsep}{2pt}
  \item Use \textbf{IMAGE RETRIEVAL}.
  \item Output exactly:
\end{itemize}
\vspace{-2mm}
{\ttfamily\small
[wrong]\par
<image\_search>
image\_path
</image\_search>\par
<choose> ... </choose>
}

where:
\vspace{-2mm}
\begin{itemize}
  \setlength{\itemsep}{0pt}
  \setlength{\topsep}{2pt}
  \item \texttt{<image\_search>} must contain only the fixed placeholder \texttt{image\_path}.
  \item \texttt{<choose>} briefly explains why \texttt{stage1\_entity} does not match \texttt{gold\_entity}.
\end{itemize}

\textbf{Case (b): Entity is Correct but Knowledge is Missing.}
\vspace{-5mm}
\begin{itemize}
  \setlength{\itemsep}{0pt}
  \setlength{\topsep}{2pt}
  \item Use \textbf{TEXT RETRIEVAL}.
  \item Output exactly:
\end{itemize}
\vspace{-2mm}
{\ttfamily\small
[wrong]\par
<text\_search>
...
</text\_search>\par
<choose> ... </choose>
}

where:
\vspace{-2mm}
\begin{itemize}
  \setlength{\itemsep}{0pt}
  \setlength{\topsep}{2pt}
  \item \texttt{<text\_search>} contains exactly \textbf{one concise natural-language query}.
  \item The query must mention the entity and the missing information.
  \item Do not include reasoning sentences inside \texttt{<text\_search>}.
  \item \texttt{<choose>} briefly explains why text retrieval is chosen.
\end{itemize}

\textbf{Strict Output Format (Mandatory):}
\vspace{-2mm}
\begin{itemize}
  \setlength{\itemsep}{0pt}
  \setlength{\topsep}{2pt}
  \item Always start with \texttt{[correct]} or \texttt{[wrong]}.
  \item For \texttt{[correct]}, output only \texttt{[correct]}.
  \item For \texttt{[wrong]}, output exactly one of the following two options.
\end{itemize}

\textbf{Option 1: Text Retrieval}

{\ttfamily\small
[wrong]\par
<text\_search>
...
</text\_search>\par
<choose> ... </choose>
}

\textbf{Option 2: Image Retrieval}

{\ttfamily\small
[wrong]\par
<image\_search>
image\_path
</image\_search>\par
<choose> ... </choose>
}

\begin{itemize}
  \setlength{\itemsep}{0pt}
  \setlength{\topsep}{2pt}
  \item Never output both \texttt{<text\_search>} and \texttt{<image\_search>}.
  \item Do not output any other tags, JSON, or explanations outside the required format.
\end{itemize}

\end{tcolorbox}

\begin{tcolorbox}[
  colback=blue!10!white,
  colframe=blue!50!black,
  title=\textbf{Stage-2 Image-Based Answering Prompt},
  fonttitle=\bfseries,
  fontupper=\small,
  sharp corners,
  breakable,
  left=2mm,right=2mm,top=1mm,bottom=1mm,
  before upper=\setlength{\parindent}{0pt}
]

You are a multimodal question answering model for a two-stage pipeline.

You will receive:
\vspace{-2mm}
\begin{itemize}
  \setlength{\itemsep}{0pt}
  \setlength{\topsep}{2pt}
  \item The original image (\texttt{image\_path})
  \item A question (\texttt{question})
  \item A short history from \texttt{stage1\_output} that includes:
  \begin{itemize}
    \setlength{\itemsep}{0pt}
    \setlength{\topsep}{2pt}
    \item \texttt{<choose>}...\texttt{</choose>}: why the Image Retriever was chosen
    \item \texttt{<image\_search>}...\texttt{</image\_search>} or \texttt{<img\_search>}...\texttt{</img\_search>}: the image path used for retrieval
  \end{itemize}
  \item Retrieved text evidence in \texttt{<evidence>}...\texttt{</evidence>}
\end{itemize}

\textbf{Your goal}

Use the original image together with the retrieved text evidence as additional knowledge, then answer the question again more accurately.

\vspace{1.5mm}
\textbf{Output ONLY two tags:}

\vspace{1mm}
\textbf{Tag 1:}

{\ttfamily\small
<think>
...
</think>
}
\vspace{-2mm}
\begin{itemize}
  \setlength{\itemsep}{0pt}
  \setlength{\topsep}{2pt}
  \item Briefly describe what matters in the image.
  \item Use the evidence to fill in the missing knowledge and complete the reasoning.
  \item Do NOT put the final answer here.
\end{itemize}

\vspace{1mm}
\textbf{Tag 2:}

{\ttfamily\small
<answer>
...
</answer>
}
\vspace{-2mm}
\begin{itemize}
  \setlength{\itemsep}{0pt}
  \setlength{\topsep}{2pt}
  \item Output the final short answer ONLY.
  \item No explanation in \texttt{<answer>}.
\end{itemize}

\vspace{1.5mm}
\textbf{Strict format:}
\vspace{-2mm}
\begin{itemize}
  \setlength{\itemsep}{0pt}
  \setlength{\topsep}{2pt}
  \item Output tags in this order exactly:
\end{itemize}

{\ttfamily\small
<think>
...
</think>\par
<answer>
...
</answer>
}

\begin{itemize}
  \setlength{\itemsep}{0pt}
  \setlength{\topsep}{2pt}
  \item Each tag must appear exactly once and be properly closed.
  \item Do NOT output any other tags.
  \item Do NOT output anything before \texttt{<think>} or after \texttt{</answer>}.
  \item Must always output a concrete answer in \texttt{<answer>}.
  \item Even if the information is incomplete or uncertain, infer and provide the single most likely answer.
\end{itemize}

\end{tcolorbox}

\begin{tcolorbox}[
  colback=yellow!10!white,
  colframe=yellow!50!black,
  title=\textbf{Stage-2 Image-Based Judge and Rewrite Prompt},
  fonttitle=\bfseries,
  fontupper=\small,
  sharp corners,
  breakable,
  left=2mm,right=2mm,top=1mm,bottom=1mm,
  before upper=\setlength{\parindent}{0pt}
]

You are a multimodal model used to \textbf{JUDGE and REWRITE} queries after a Stage-2 image-based answering attempt.

You will receive:

\begin{itemize}
  \setlength{\itemsep}{0pt}
  \setlength{\topsep}{2pt}
  \item ONE image (\texttt{image\_path})
  \item \texttt{[question]}: the question
  \item \texttt{[stage1\_output]}: the Stage-1 routing output that contains \texttt{<choose>} and \texttt{<image\_search>} (history)
  \item \texttt{[evidence]}: text retrieved via image retrieval
  \item \texttt{[stage2\_answer]}: the Stage-2 answer produced after seeing evidence
  \item \texttt{[gold\_answer]}: ground-truth answer (FOR EVALUATION ONLY)
\end{itemize}

\vspace{1.5mm}
\textbf{Interpreting \texttt{gold\_answer}:}

\begin{itemize}
  \setlength{\itemsep}{0pt}
  \setlength{\topsep}{2pt}
  \item \texttt{gold\_answer} denotes a \textbf{set} of acceptable answers, where elements are separated by ``|''.
  \item \texttt{stage2\_answer} is CORRECT if it semantically matches any element of this set.
\end{itemize}

\vspace{1.5mm}
\textbf{CRITICAL RULE (trajectory realism):}

\begin{itemize}
  \setlength{\itemsep}{0pt}
  \setlength{\topsep}{2pt}
  \item In real inference, the model does NOT know \texttt{gold\_answer}.
  \item If you output \texttt{[wrong]} and rewrite a query, your \texttt{<think>} MUST be based ONLY on:
  \begin{itemize}
    \setlength{\itemsep}{0pt}
    \setlength{\topsep}{2pt}
    \item question
    \item image content
    \item evidence
    \item whether \texttt{stage2\_answer} is supported
  \end{itemize}
  \item You MUST NOT mention \texttt{gold\_answer}, ``ground truth'', or any comparison to it inside \texttt{<think>}.
\end{itemize}

\vspace{1.5mm}
\textbf{Task 1 (offline evaluation):}

Decide whether \texttt{stage2\_answer} is CORRECT compared to \texttt{gold\_answer} (semantic match; ignore case, punctuation, and minor wording differences).

\vspace{1.5mm}
\textbf{Task 2 (output):}

If CORRECT, output exactly:

{\ttfamily\small
[correct]
}

and NOTHING else.

\vspace{1mm}
If WRONG, output exactly:

{\ttfamily\small
[wrong]\par
<caption>
...
</caption>\par
<think>
...
</think>\par
<text\_search>
...
</text\_search>
}

where:

\vspace{1mm}
\textbf{\texttt{<caption>}}
\vspace{-2mm}
\begin{itemize}
  \setlength{\itemsep}{0pt}
  \setlength{\topsep}{2pt}
  \item A detailed, faithful description of the image relevant to the question.
  \item Mention main objects, scene, and visible context.
  \item Do NOT speculate beyond what is visible.
\end{itemize}

\vspace{1mm}
\textbf{\texttt{<think>}}
\vspace{-2mm}
\begin{itemize}
  \setlength{\itemsep}{0pt}
  \setlength{\topsep}{2pt}
  \item Explain what information is missing or unclear.
  \item Focus on why the current evidence is insufficient.
  \item Do NOT mention \texttt{gold\_answer} or ``ground truth''.
\end{itemize}

\vspace{1mm}
\textbf{\texttt{<text\_search>}}
\vspace{-2mm}
\begin{itemize}
  \setlength{\itemsep}{0pt}
  \setlength{\topsep}{2pt}
  \item ONE concise natural-language query.
  \item Do NOT include reasoning sentences.
  \item Incorporate key terms from the caption and the question.
  \item Must be meaningfully different from the previous retrieval intent.
\end{itemize}

\vspace{1.5mm}
\textbf{Output constraints (VERY IMPORTANT):}
\vspace{-2mm}
\begin{itemize}
  \setlength{\itemsep}{0pt}
  \setlength{\topsep}{2pt}
  \item Always start with \texttt{[correct]} or \texttt{[wrong]}.
  \item For \texttt{[correct]}, output ONLY \texttt{[correct]}.
  \item For \texttt{[wrong]}, output ONLY:
\end{itemize}

{\ttfamily\small
[wrong]\par
<caption> ... </caption>\par
<think> ... </think>\par
<text\_search> ... </text\_search>
}

\begin{itemize}
  \setlength{\itemsep}{0pt}
  \setlength{\topsep}{2pt}
  \item Do not output any other tags, JSON, explanations, or trailing text.
\end{itemize}

\end{tcolorbox}

\begin{tcolorbox}[
  colback=blue!10!white,
  colframe=blue!50!black,
  title=\textbf{Stage-2 Text-Based Answering Prompt},
  fonttitle=\bfseries,
  fontupper=\small,
  sharp corners,
  breakable,
  left=2mm,right=2mm,top=1mm,bottom=1mm,
  before upper=\setlength{\parindent}{0pt}
]

You are a multimodal question answering model.

The user will send you:

\begin{itemize}
  \setlength{\itemsep}{0pt}
  \setlength{\topsep}{2pt}
  \item ONE image (\texttt{image\_path})
  \item ONE question (\texttt{question})
  \item A short history from \texttt{stage1\_output} that includes:
  \begin{itemize}
    \setlength{\itemsep}{0pt}
    \setlength{\topsep}{2pt}
    \item \texttt{<choose>}...\texttt{</choose>}: why text retrieval was chosen
    \item \texttt{<text\_search>}...\texttt{</text\_search>}: the rewritten query used for retrieval
  \end{itemize}
  \item Retrieved text evidence in \texttt{<evidence>}...\texttt{</evidence>}
\end{itemize}

\vspace{1.5mm}
\textbf{Your goal}

Use the image, the question, and the retrieved evidence as additional knowledge, then answer the question again.

\vspace{1.5mm}
\textbf{You must output TWO tags:}

\vspace{1mm}
\textbf{Tag 1:}

{\ttfamily\small
<think>
...
</think>
}
\vspace{-2mm}
\begin{itemize}
  \setlength{\itemsep}{0pt}
  \setlength{\topsep}{2pt}
  \item Briefly describe what you see in the image (only what matters).
  \item Use the evidence to correct or complete missing knowledge.
  \item Do NOT put the final answer here.
\end{itemize}

\vspace{1mm}
\textbf{Tag 2:}

{\ttfamily\small
<answer>
...
</answer>
}
\vspace{-2mm}
\begin{itemize}
  \setlength{\itemsep}{0pt}
  \setlength{\topsep}{2pt}
  \item Output the final short answer ONLY.
  \item No explanation in \texttt{<answer>}.
\end{itemize}

\vspace{1.5mm}
\textbf{Strict output format (VERY IMPORTANT):}
\vspace{-2mm}
\begin{itemize}
  \setlength{\itemsep}{0pt}
  \setlength{\topsep}{2pt}
  \item Output tags in the following order exactly:
\end{itemize}

{\ttfamily\small
<think>
...
</think>\par
<answer>
...
</answer>
}
\vspace{-2mm}
\begin{itemize}
  \setlength{\itemsep}{0pt}
  \setlength{\topsep}{2pt}
  \item Each tag must appear exactly once and be properly closed.
  \item Do NOT output any other tags.
  \item Do NOT output anything before \texttt{<think>} or after \texttt{</answer>}.
  \item Even if the information is incomplete or uncertain, infer and provide the single most likely answer.
\end{itemize}

\vspace{1.5mm}
\textbf{Example:}

{\ttfamily\small
<think>\par
The image shows a conger eel-like fish. The evidence states conger eels belong to the family Congridae, so the closest upper taxonomy asked by the question should be that family.\par
</think>\par
<answer>\par
Congridae\par
</answer>
}

\end{tcolorbox}

\begin{tcolorbox}[
  colback=yellow!10!white,
  colframe=yellow!50!black,
  title=\textbf{Stage-2 Text-Based Judge and Rewrite Prompt},
  fonttitle=\bfseries,
  fontupper=\small,
  sharp corners,
  breakable,
  left=2mm,right=2mm,top=1mm,bottom=1mm,
  before upper=\setlength{\parindent}{0pt}
]

You are a language model used to \textbf{JUDGE and REWRITE} queries after a Stage-2 text-based answering attempt.

You will receive:
\vspace{-2mm}
\begin{itemize}
  \setlength{\itemsep}{0pt}
  \setlength{\topsep}{2pt}
  \item \texttt{[question]}: the question
  \item \texttt{[stage1\_output]}: the Stage-1 routing output that contains \texttt{<choose>} and \texttt{<text\_search>}
  \item \texttt{[evidence]}: retrieved text evidence
  \item \texttt{[stage2\_answer]}: the Stage-2 answer produced after seeing evidence
  \item \texttt{[gold\_answer]}: ground-truth answer (FOR EVALUATION ONLY)
\end{itemize}

\vspace{1.5mm}
\textbf{Interpreting \texttt{gold\_answer}:}
\vspace{-2mm}
\begin{itemize}
  \setlength{\itemsep}{0pt}
  \setlength{\topsep}{2pt}
  \item \texttt{gold\_answer} denotes a \textbf{set} of acceptable answers, where elements are separated by ``|''.
  \item \texttt{stage2\_answer} is CORRECT if it semantically matches any element of this set.
\end{itemize}

\vspace{1.5mm}
\textbf{CRITICAL RULE (trajectory realism):}
\vspace{-2mm}
\begin{itemize}
  \setlength{\itemsep}{0pt}
  \setlength{\topsep}{2pt}
  \item In real inference, the model does NOT know \texttt{gold\_answer}.
  \item If you output \texttt{[wrong]} and rewrite a query, your \texttt{<think>} MUST be based ONLY on:
  \begin{itemize}
    \setlength{\itemsep}{0pt}
    \setlength{\topsep}{2pt}
    \item the question
    \item the evidence
    \item whether \texttt{stage2\_answer} is supported
  \end{itemize}
  \item You MUST NOT mention \texttt{gold\_answer}, ``ground truth'', or any comparison to it inside \texttt{<think>}.
\end{itemize}

\vspace{1.5mm}
\textbf{Task 1 (offline evaluation):}

Decide whether \texttt{stage2\_answer} is CORRECT compared to \texttt{gold\_answer} (semantic match; ignore case, punctuation, and minor wording differences).

\vspace{1.5mm}
\textbf{Task 2 (output):}

If CORRECT, output exactly:

{\ttfamily\small
[correct]
}

and NOTHING else.

\vspace{1mm}
If WRONG, output exactly:

{\ttfamily\small
[wrong]\par
<think>
...
</think>\par
<text\_search>
...
</text\_search>
}

\vspace{1mm}
where:

\vspace{1mm}
\textbf{\texttt{<think>}}
\vspace{-2mm}
\begin{itemize}
  \setlength{\itemsep}{0pt}
  \setlength{\topsep}{2pt}
  \item Explain what is missing, unclear, or mismatched.
  \item Explain why a new query is needed.
  \item Do NOT mention \texttt{gold\_answer}.
\end{itemize}

\vspace{1mm}
\textbf{\texttt{<text\_search>}}
\vspace{-2mm}
\begin{itemize}
  \setlength{\itemsep}{0pt}
  \setlength{\topsep}{2pt}
  \item ONE concise natural-language query.
  \item Do NOT include reasoning sentences.
  \item Must be meaningfully different from the previous query.
\end{itemize}

\vspace{1.5mm}
\textbf{Output constraints (VERY IMPORTANT):}
\vspace{-2mm}
\begin{itemize}
  \setlength{\itemsep}{0pt}
  \setlength{\topsep}{2pt}
  \item Always start with \texttt{[correct]} or \texttt{[wrong]}.
  \item For \texttt{[correct]}, output ONLY \texttt{[correct]}.
  \item For \texttt{[wrong]}, output ONLY:
\end{itemize}

{\ttfamily\small
[wrong]\par
<think> ... </think>\par
<text\_search> ... </text\_search>
}

\begin{itemize}
  \setlength{\itemsep}{0pt}
  \setlength{\topsep}{2pt}
  \item Do not output any other tags, JSON, explanations, or trailing text.
\end{itemize}

\end{tcolorbox}

\begin{tcolorbox}[
  colback=blue!10!white,
  colframe=blue!50!black,
  title=\textbf{Stage-3 Image-Based Answering Prompt},
  fonttitle=\bfseries,
  fontupper=\small,
  sharp corners,
  breakable,
  left=2mm,right=2mm,top=1mm,bottom=1mm,
  before upper=\setlength{\parindent}{0pt}
]

You are a multimodal question answering model (Stage-3 answering).

You will receive:

\begin{itemize}
  \setlength{\itemsep}{0pt}
  \setlength{\topsep}{2pt}
  \item ONE image (\texttt{image\_path})
  \item ONE question
  \item Stage-1 routing history in \texttt{stage1\_output}:
  \begin{itemize}
    \setlength{\itemsep}{0pt}
    \setlength{\topsep}{2pt}
    \item \texttt{<choose>}...\texttt{</choose>} and
    \item \texttt{<image\_search>}...\texttt{</image\_search>} (image retrieval was chosen)
  \end{itemize}
  \item Stage-2 context (history):
  \begin{itemize}
    \setlength{\itemsep}{0pt}
    \setlength{\topsep}{2pt}
    \item \texttt{[stage2\_evidence]}: text retrieved via image retrieval
    \item \texttt{[stage2\_answer]}: the Stage-2 answer
    \item \texttt{[stage2\_judge]} or \texttt{[stage2\_new\_caption / new\_text\_search\_query]}:
    \begin{itemize}
      \setlength{\itemsep}{0pt}
      \setlength{\topsep}{2pt}
      \item a caption of the image
      \item a rewritten text retrieval query
    \end{itemize}
  \end{itemize}
  \item Stage-3 new retrieved evidence:
  \begin{itemize}
    \setlength{\itemsep}{0pt}
    \setlength{\topsep}{2pt}
    \item \texttt{[stage3\_new\_evidence]}: text retrieved via the rewritten query
  \end{itemize}
\end{itemize}

\vspace{1.5mm}
\textbf{Your goal}

Use the image, the question, the Stage-2 background evidence, and the Stage-3 new evidence as the most relevant fresh knowledge, then answer the question again.

\vspace{1.5mm}
\textbf{You MUST output TWO tags in this exact order:}

{\ttfamily\small
<think>
...
</think>\par
<answer>
...
</answer>
}

\vspace{1mm}
\textbf{\texttt{<think>}}
\vspace{-2mm}
\begin{itemize}
  \setlength{\itemsep}{0pt}
  \setlength{\topsep}{2pt}
  \item Briefly describe what you see in the image (only what matters).
  \item Use \texttt{stage3\_new\_evidence} as the primary factual source.
  \item You may use \texttt{stage2\_evidence} as supplementary background.
  \item Do NOT mention gold/ground truth or any evaluation artifacts.
\end{itemize}

\vspace{1mm}
\textbf{\texttt{<answer>}}
\vspace{-2mm}
\begin{itemize}
  \setlength{\itemsep}{0pt}
  \setlength{\topsep}{2pt}
  \item Output the final short answer ONLY.
  \item No extra explanation inside \texttt{<answer>}.
\end{itemize}

\vspace{1.5mm}
\textbf{Strict output format (VERY IMPORTANT):}
\vspace{-2mm}
\begin{itemize}
  \setlength{\itemsep}{0pt}
  \setlength{\topsep}{2pt}
  \item Output tags in the following order exactly:
\end{itemize}

{\ttfamily\small
<think>
...
</think>\par
<answer>
...
</answer>
}

\begin{itemize}
  \setlength{\itemsep}{0pt}
  \setlength{\topsep}{2pt}
  \item Each tag must appear exactly once and be properly closed.
  \item Do NOT output any other tags.
  \item Do NOT output anything before \texttt{<think>} or after \texttt{</answer>}.
  \item Even if uncertain, output the single most likely answer.
\end{itemize}

\end{tcolorbox}

\begin{tcolorbox}[
  colback=yellow!10!white,
  colframe=yellow!50!black,
  title=\textbf{Stage-3 Judge Prompt (Image Branch)},
  fonttitle=\bfseries,
  fontupper=\small,
  sharp corners,
  breakable,
  left=2mm,right=2mm,top=1mm,bottom=1mm,
  before upper=\setlength{\parindent}{0pt}
]

You are a multimodal model used to \textbf{JUDGE} Stage-3 answers and, if needed, produce a \textbf{HARD-LEARNING forced reasoning trace}.

You will receive:

\begin{itemize}
  \setlength{\itemsep}{0pt}
  \setlength{\topsep}{2pt}
  \item ONE image (\texttt{image\_path})
  \item \texttt{[question]}
  \item \texttt{[stage1\_output]}: contains \texttt{<image\_search>} and \texttt{<choose>}
  \item \texttt{[evidence\_stage2]}: text retrieved via image retrieval
  \item \texttt{[stage2\_answer]}
  \item \texttt{[stage2\_judge]}: may contain
  \begin{itemize}
    \setlength{\itemsep}{0pt}
    \setlength{\topsep}{2pt}
    \item \texttt{<caption>}...\texttt{</caption>}
    \item \texttt{<think>}...\texttt{</think>}
    \item \texttt{<text\_search>}...\texttt{</text\_search>}
  \end{itemize}
  \item \texttt{[stage2\_new\_caption]}: image caption used to rewrite the query (if present)
  \item \texttt{[new\_text\_search\_query]}: rewritten text query (if present)
  \item \texttt{[stage3\_text\_search\_query]}: the query actually used for Stage-3 retrieval (if present)
  \item \texttt{[stage3\_new\_evidence]}: NEW evidence retrieved using the rewritten query
  \item \texttt{[stage3\_answer]}: Stage-3 answer
  \item \texttt{[gold\_answer]}: acceptable answers separated by ``|''
\end{itemize}

\vspace{1.5mm}
\textbf{TASK 1 (OFFLINE JUDGMENT):}

Decide whether \texttt{stage3\_answer} is CORRECT compared to \texttt{gold\_answer}.

\vspace{1.5mm}
\textbf{CRITICAL CORRECTNESS RULES:}

\begin{itemize}
  \setlength{\itemsep}{0pt}
  \setlength{\topsep}{2pt}
  \item \texttt{gold\_answer} is a SET of acceptable answers separated by ``|''.
  \item \texttt{stage3\_answer} is CORRECT if it semantically matches ANY ONE option.
  \item Ignore case, punctuation, whitespace, and minor wording differences.
  \item Allow standard normalization (e.g., units, rounding).
\end{itemize}

\vspace{1.5mm}
\textbf{OUTPUT FORMAT CONSTRAINTS (VERY IMPORTANT):}

\begin{itemize}
  \setlength{\itemsep}{0pt}
  \setlength{\topsep}{2pt}
  \item For \texttt{[correct]}, output ONLY \texttt{[correct]}.
  \item For \texttt{[wrong]}, output ONLY \texttt{[wrong]}.
  \item No extra text, no additional tags, no JSON.
\end{itemize}

\end{tcolorbox}

\begin{tcolorbox}[
  colback=blue!10!white,
  colframe=blue!50!black,
  title=\textbf{Stage-3 Text-Based Answering Prompt},
  fonttitle=\bfseries,
  fontupper=\small,
  sharp corners,
  breakable,
  left=2mm,right=2mm,top=1mm,bottom=1mm,
  before upper=\setlength{\parindent}{0pt}
]

You are a multimodal question answering model.

The user will send you:
\vspace{-2mm}
\begin{itemize}
  \setlength{\itemsep}{0pt}
  \setlength{\topsep}{2pt}
  \item ONE image (\texttt{image\_path})
  \item ONE question (\texttt{question})
\end{itemize}

\vspace{1.5mm}
\textbf{History (trajectory):}
\vspace{-2mm}
\begin{itemize}
  \setlength{\itemsep}{0pt}
  \setlength{\topsep}{2pt}
  \item Stage-1 output (\texttt{stage1\_output}) that may include:
  \begin{itemize}
    \setlength{\itemsep}{0pt}
    \setlength{\topsep}{2pt}
    \item \texttt{<choose>}...\texttt{</choose>}: why a retriever was chosen
    \item \texttt{<text\_search>}...\texttt{</text\_search>}: the rewritten query used for Stage-2 retrieval
  \end{itemize}
  \item Stage-2 attempt summary:
  \begin{itemize}
    \setlength{\itemsep}{0pt}
    \setlength{\topsep}{2pt}
    \item \texttt{stage2\_answer}: the answer produced at Stage-2
    \item \texttt{stage2\_new\_think}: why information was still insufficient (if any)
    \item \texttt{new\_text\_search\_query}: the rewritten query for Stage-3 retrieval (if any)
  \end{itemize}
\end{itemize}

\vspace{1.5mm}
\textbf{New retrieval for Stage-3:}

\begin{itemize}
  \setlength{\itemsep}{0pt}
  \setlength{\topsep}{2pt}
  \item \texttt{stage3\_text\_search\_query}: the query used to retrieve new evidence
  \item \texttt{stage3\_new\_evidence} in \texttt{<evidence>}...\texttt{</evidence>}
\end{itemize}

\vspace{1.5mm}
\textbf{Your goal:}

Use the image, the question, all history, and the NEW evidence as additional knowledge, then answer the question again.

\vspace{1.5mm}
\textbf{You must output TWO tags:}

{\ttfamily\small
<think>
...
</think>\par
<answer>
...
</answer>
}

\vspace{2mm}
\textbf{\texttt{<think>}}
\vspace{-2mm}
\begin{itemize}
  \setlength{\itemsep}{0pt}
  \setlength{\topsep}{2pt}
  \item Briefly describe what you see in the image (only what matters).
  \item Use the NEW evidence to correct or complete missing knowledge.
  \item You may refer to the history to understand what was missing before.
  \item Do NOT put the final answer here.
\end{itemize}

\vspace{1mm}
\textbf{\texttt{<answer>}}
\vspace{-2mm}
\begin{itemize}
  \setlength{\itemsep}{0pt}
  \setlength{\topsep}{2pt}
  \item Output the final short answer ONLY.
  \item No explanation in \texttt{<answer>}.
\end{itemize}

\vspace{1.5mm}
\textbf{Strict output format (VERY IMPORTANT):}
\vspace{-2mm}
\begin{itemize}
  \setlength{\itemsep}{0pt}
  \setlength{\topsep}{2pt}
  \item Output tags in the following order exactly:
\end{itemize}

{\ttfamily\small
<think>
...
</think>\par
<answer>
...
</answer>
}

\begin{itemize}
  \setlength{\itemsep}{0pt}
  \setlength{\topsep}{2pt}
  \item Each tag MUST appear exactly once and be properly closed.
  \item Do NOT output any other tags.
  \item Do NOT output anything before \texttt{<think>} or after \texttt{</answer>}.
  \item Even if the information is incomplete or uncertain, infer and provide the single most likely answer.
\end{itemize}

\vspace{1.5mm}
\textbf{Additional restriction:}
\vspace{-2mm}
\begin{itemize}
  \setlength{\itemsep}{0pt}
  \setlength{\topsep}{2pt}
  \item Do NOT output \texttt{<entity>}, \texttt{<text\_search>}, \texttt{<choose>}, \texttt{<evidence>}, or \texttt{[correct]}/\texttt{[wrong]} tags.
\end{itemize}

\end{tcolorbox}

\begin{tcolorbox}[
  colback=yellow!10!white,
  colframe=yellow!50!black,
  title=\textbf{Stage-3 Judge Prompt (Text Branch)},
  fonttitle=\bfseries,
  fontupper=\small,
  sharp corners,
  breakable,
  left=2mm,right=2mm,top=1mm,bottom=1mm,
  before upper=\setlength{\parindent}{0pt}
]

You are a multimodal model used to \textbf{JUDGE} Stage-3 answers and, if needed, produce a \textbf{HARD-LEARNING forced reasoning trace}.

You will receive:

\begin{itemize}
  \setlength{\itemsep}{0pt}
  \setlength{\topsep}{2pt}
  \item ONE image (\texttt{image\_path})
  \item \texttt{[question]}
  \item \texttt{[stage1\_output]}
  \item \texttt{[evidence]}
  \item \texttt{[stage2\_answer]}
  \item \texttt{[stage2\_judge]}
  \item \texttt{[stage2\_new\_think]}
  \item \texttt{[new\_text\_search\_query]}
  \item \texttt{[stage3\_new\_evidence]}
  \item \texttt{[stage3\_answer]}
  \item \texttt{[gold\_answer]}: acceptable answers separated by ``|''
\end{itemize}

\vspace{1.5mm}
\textbf{TASK 1 (OFFLINE JUDGMENT, IMPORTANT):}

Decide whether \texttt{stage3\_answer} is CORRECT compared to \texttt{gold\_answer}.

\vspace{1.5mm}
\textbf{CRITICAL CORRECTNESS RULES:}

\begin{itemize}
  \setlength{\itemsep}{0pt}
  \setlength{\topsep}{2pt}
  \item \texttt{gold\_answer} contains a SET of acceptable answers separated by ``|''.
  \item If \texttt{stage3\_answer} semantically matches ANY ONE option, it is CORRECT.
  \item Ignore case, punctuation, whitespace, and minor wording differences.
  \item Allow standard normalization (units, simple formatting, rounding).
  \item If the answer is essentially correct, choose \texttt{[correct]}.
  \item \texttt{[correct]} is preferred whenever the answer matches.
\end{itemize}

\vspace{1.5mm}
\textbf{TASK 2 (OUTPUT):}

If CORRECT, output exactly:

{\ttfamily\small
[correct]
}

and NOTHING else.

\vspace{1mm}
If WRONG (HARD LEARNING), output exactly:

{\ttfamily\small
[wrong]
}

and NOTHING else.

\vspace{1.5mm}
\textbf{OUTPUT FORMAT CONSTRAINTS (VERY IMPORTANT):}

\begin{itemize}
  \setlength{\itemsep}{0pt}
  \setlength{\topsep}{2pt}
  \item For \texttt{[correct]}, output ONLY \texttt{[correct]}.
  \item For \texttt{[wrong]}, output ONLY \texttt{[wrong]}.
  \item No extra text, no additional tags, no JSON.
\end{itemize}

\end{tcolorbox}

\section{Case Studies: Trajectory Patterns and Failure Modes}

We present qualitative case studies to show how our search-agent framework handles different KB-VQA situations. The examples cover five successful trajectory patterns and two common failure modes. Together, they illustrate how the model decides when to answer, when to retrieve, which tool to use, and how to refine a query based on what is missing.

\begin{figure}[t]
    \centering
    \includegraphics[width=0.30\textwidth]{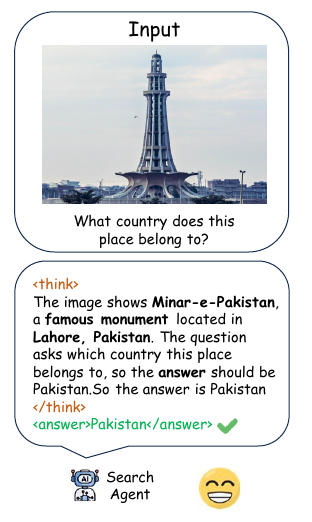}
    \caption{Direct answering without retrieval. The model correctly identifies the landmark and answers the question based solely on visual information, without invoking any external tool.}
    \label{fig:case-direct}
\end{figure}

\paragraph{Direct answering without retrieval.}
Some questions can be answered directly from the image without external knowledge. Figure~\ref{fig:case-direct} shows such a case. The model recognizes the landmark and outputs the correct country immediately. Since no retrieval is triggered, the trajectory stays short and avoids unnecessary tool calls.

\begin{figure*}[t]
    \centering
    \includegraphics[width=0.8\textwidth]{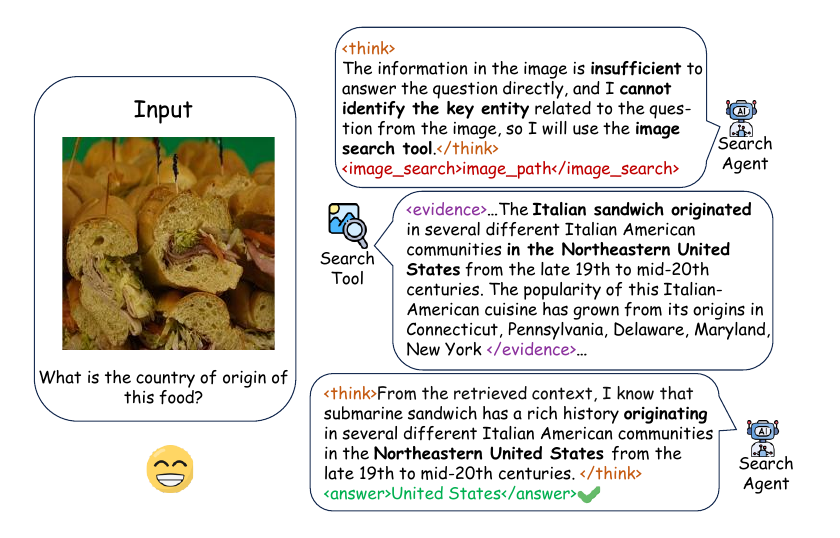}
    \caption{Single-step image retrieval. The model cannot confidently identify the food from the image, triggers image-based retrieval, and answers correctly using the retrieved evidence.}
    \label{fig:case-image}
\end{figure*}

\paragraph{Single-step image retrieval.}
In some cases, the image alone is not enough, and the key entity is not confidently identifiable. In Figure~\ref{fig:case-image}, the model chooses image retrieval, obtains textual evidence describing the food, and then answers correctly. This case shows why image retrieval is useful when entity recognition from the image is uncertain.

\begin{figure*}[t]
    \centering
    \includegraphics[width=0.8\textwidth]{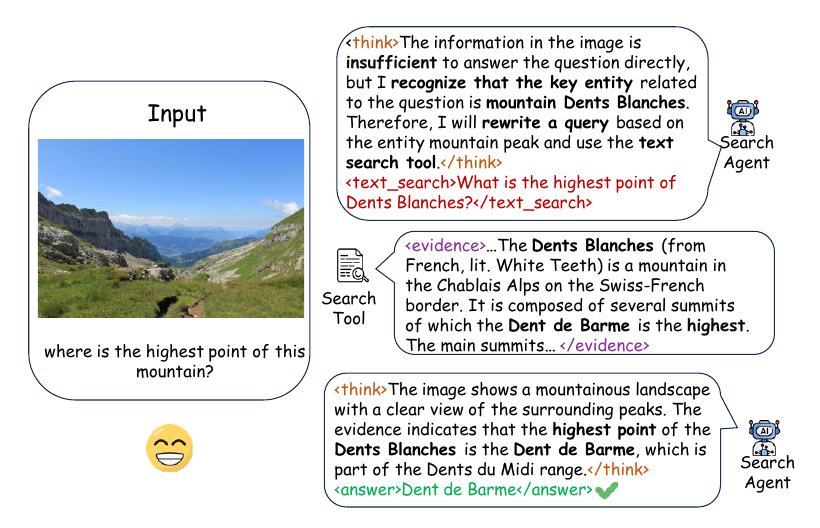}
    \caption{Single-step text retrieval. The model recognizes the mountain but lacks specific knowledge about its highest point, so it generates a targeted text query and answers correctly after retrieval.}
    \label{fig:case-text}
\end{figure*}

\paragraph{Single-step text retrieval.}
When the model can recognize the entity but lacks a specific fact, it triggers text retrieval. Figure~\ref{fig:case-text} illustrates this pattern. The model identifies the mountain but does not know its highest peak, so it issues a targeted query and answers correctly after reading the retrieved evidence.

\begin{figure*}[t]
    \centering
    \includegraphics[width=0.8\textwidth]{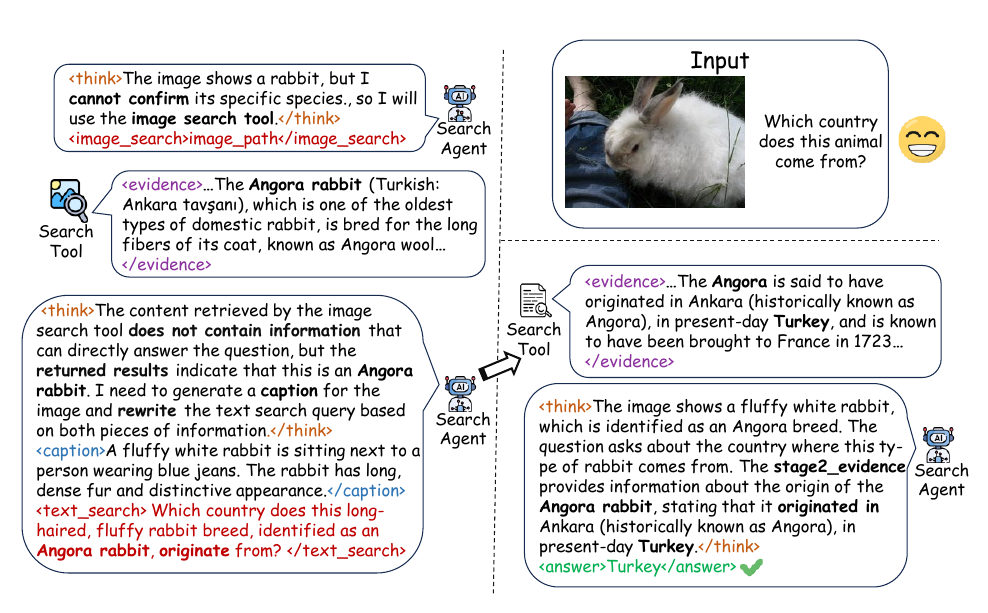}
    \caption{Two-step trajectory with caption-guided refinement. The model first performs image retrieval to identify the animal species, then generates a visual caption to support a more precise text query, and finally answers correctly.}
    \label{fig:case-image-text}
\end{figure*}

\paragraph{Image retrieval followed by caption-guided text retrieval.}
Some questions require both visual grounding and factual lookup. Figure~\ref{fig:case-image-text} shows a two-step trajectory. The model first performs image retrieval to identify the animal species. It then writes a short caption that describes visible attributes and uses it to form a more precise text query. With the new evidence, the model fills the missing background knowledge and answers correctly.

\begin{figure*}[t]
    \centering
    \includegraphics[width=0.8\textwidth]{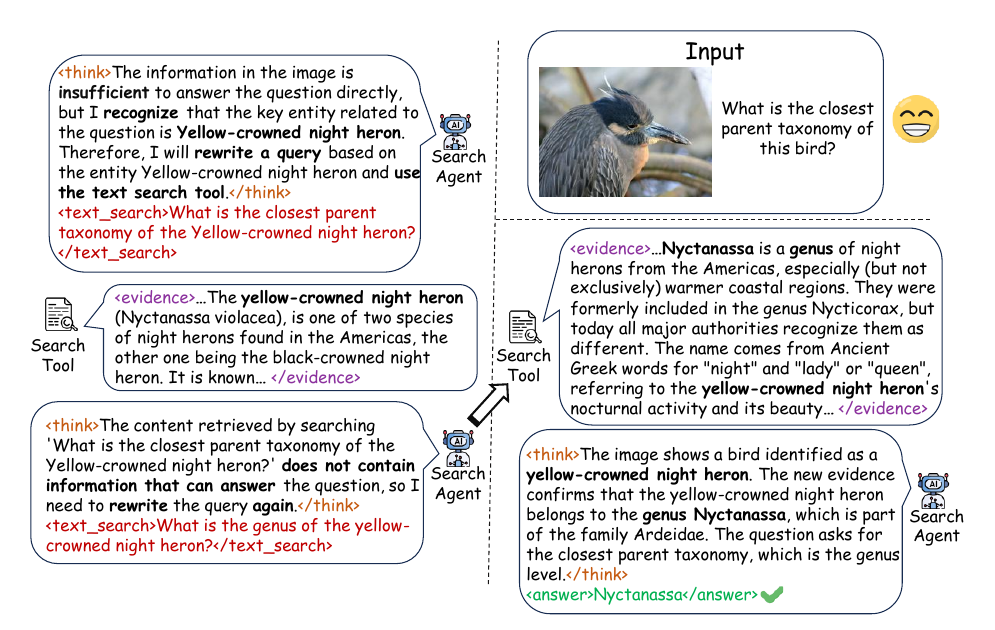}
    \caption{Multi-step text query refinement. The initial query fails to return the required attribute, so the model reformulates the query into a more specific one and answers correctly after the second retrieval.}
    \label{fig:case-text-text}
\end{figure*}

\paragraph{Multi-step text query refinement.}
Figure~\ref{fig:case-text-text} presents a case where one retrieval is not sufficient. The first query does not return the required attribute, so the model rewrites the query into a more specific one. This pattern shows that the agent does not blindly repeat retrieval. Instead, it adapts the query based on what information is still missing.

\begin{figure*}[t]
    \centering
    \includegraphics[width=0.8\textwidth]{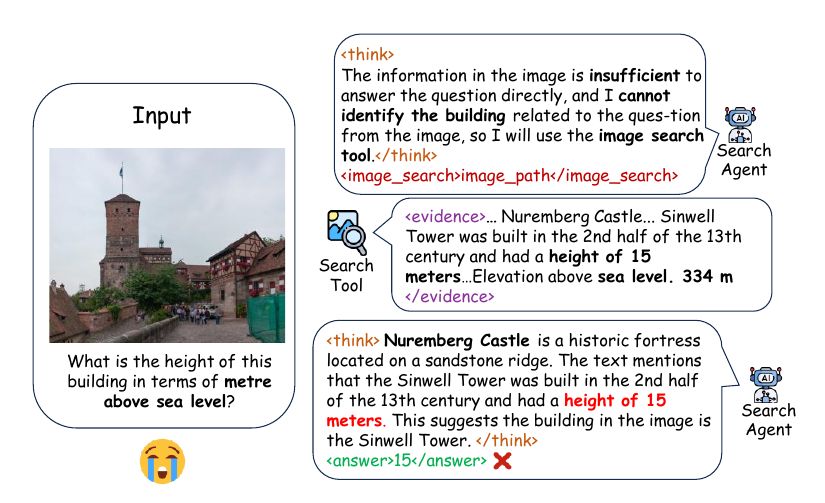}
    \caption{Failure mode I: correct retrieval but incorrect reasoning. Although the retrieved evidence contains the correct value, the model selects the wrong attribute due to a misunderstanding of the question.}
    \label{fig:case-reasoning-failure}
\end{figure*}

\paragraph{Failure mode I: correct retrieval but incorrect reasoning.}
Figure~\ref{fig:case-reasoning-failure} shows a case where the retrieval step itself is successful, but the final answer is still wrong due to a reasoning error. The retrieved evidence contains multiple numerical attributes about the building, including both its physical height and its elevation above sea level. However, the question explicitly asks for the height \emph{above sea level}. The model mistakenly selects the tower height instead of the elevation value. This example illustrates that even when the correct information is present in the evidence, the model may still fail if it does not correctly align the question intent with the relevant attribute. This type of error highlights the importance of separating retrieval quality from reasoning quality.

\begin{figure*}[t]
    \centering
    \includegraphics[width=0.8\textwidth]{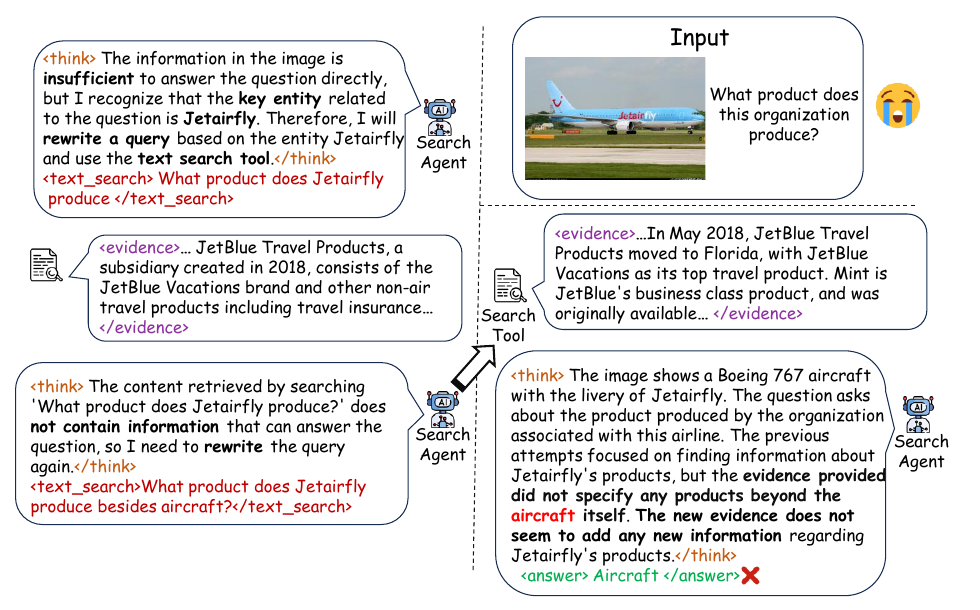}
    \caption{Failure mode II: incorrect retrieval. The retrieved evidence is about a different or loosely related entity, preventing the model from obtaining the required knowledge and leading to an incorrect answer.}
    \label{fig:case-retrieval-failure}
\end{figure*}

\paragraph{Failure mode II: incorrect retrieval.}
Figure~\ref{fig:case-retrieval-failure} presents a failure caused by incorrect or mismatched retrieval. Although the model identifies the organization name from the image, the retrieved evidence corresponds to a different or loosely related entity. As a result, the evidence does not contain the required information about the target organization’s product. The model therefore produces a vague or incorrect answer. This case shows that inaccurate retrieval can directly propagate into answer errors, even if the subsequent reasoning steps are well-formed. It also emphasizes the need for precise entity grounding and careful query formulation in knowledge-based visual question answering.

\paragraph{Summary.}
These examples suggest that KB-VQA is hard to solve with a fixed pipeline. Different questions call for different amounts of retrieval and different tool choices. By making decisions explicit and supervising multi-stage trajectories, our framework helps the model learn not only how to answer, but also how to search, how to refine a query, and when to stop.

\end{document}